\documentclass[10pt,twocolumn,letterpaper]{article}

\usepackage{cvpr}              %

\usepackage{tabularx}
\usepackage[export]{adjustbox}
\usepackage{multirow}
\usepackage{comment}

\newcolumntype{Y}{>{\centering\arraybackslash}X}
\newcolumntype{P}[1]{>{\centering\arraybackslash}p{#1}}
\newcolumntype{C}{>{\centering}X}

\newcommand{\Ours}{FlexAvatar}
\newcommand{\flextokens}{bias sinks}

\definecolor{cvprblue}{rgb}{0.21,0.49,0.74}
\usepackage[pagebackref,breaklinks,colorlinks,allcolors=cvprblue]{hyperref}

\def\paperID{40548} %
\def\confName{CVPR}
\def\confYear{2025}

\title{FlexAvatar: Learning Complete 3D Head Avatars with Partial Supervision}

\author{Tobias Kirschstein$^{1}$ \, Simon Giebenhain$^{1}$ \, Matthias Nießner$^{1}$ \\
  Technical University of Munich$^{1}$}

\begin{document}
\twocolumn[{
\renewcommand\twocolumn[1][]{#1}%
\maketitle
\vspace{-0.5cm}
\includegraphics[width=\textwidth]{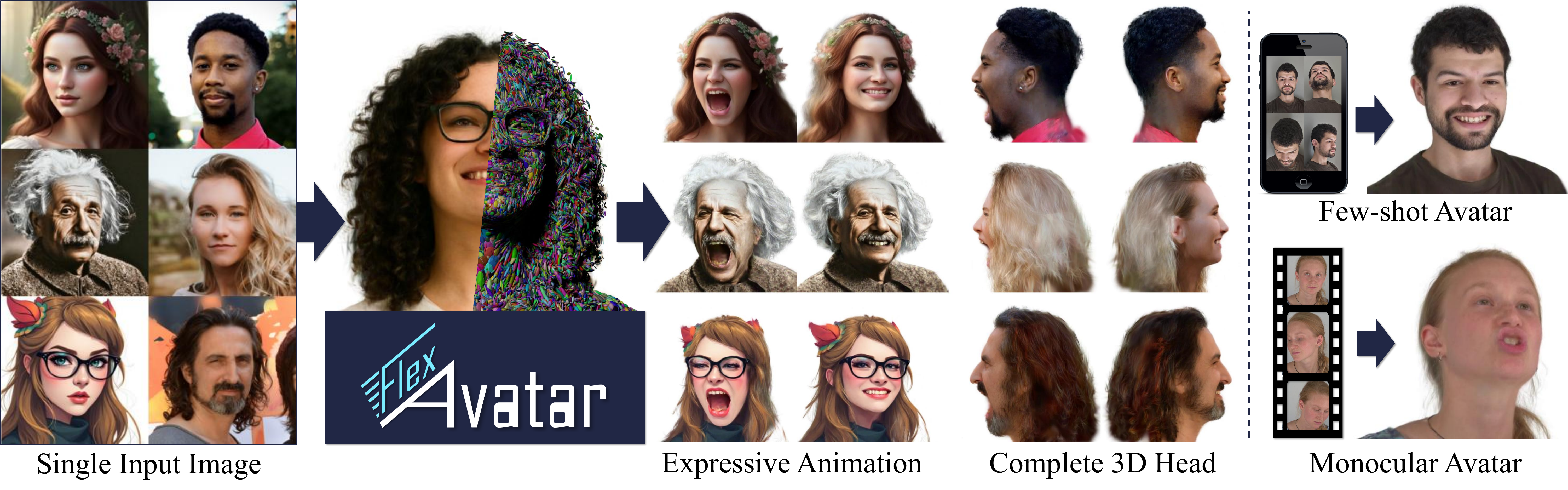}
\vspace{-0.7cm}
\captionof{figure}{\textbf{{\Ours}.} From just a single portrait image of a person, {\Ours} creates a high quality 3D head avatar representation that can be freely animated and rendered from diverse viewpoints. Our model can be flexibly applied to other scenarios including creating avatars from a phone scan or from monocular videos. The entire avatar creation process can be executed within minutes.} \vspace{2em}
\label{fig:teaser}
}]

\maketitle
\begin{abstract}
We introduce {\Ours}, a method for creating high-quality and complete 3D head avatars from a single image. 
A core challenge lies in the limited availability of multi-view data and the tendency of monocular training to yield incomplete 3D head reconstructions.
We identify the root cause of this issue as the entanglement between driving signal and target viewpoint when learning from monocular videos.
To address this, we propose a transformer-based 3D portrait animation model with learnable data source tokens, so-called {\flextokens}, which enables unified training across monocular and multi-view datasets. This design leverages the strengths of both data sources during inference: strong generalization from monocular data and full 3D completeness from multi-view supervision.
Furthermore, our training procedure yields a smooth latent avatar space that facilitates identity interpolation and flexible fitting to an arbitrary number of input observations.
In extensive evaluations on single-view, few-shot, and monocular avatar creation tasks, we verify the efficacy of {\Ours}. Many existing methods struggle with view extrapolation while {\Ours} generates complete 3D head avatars with realistic facial animations.
\\
\urlstyle{same}Website: {\url{https://tobias-kirschstein.github.io/flexavatar/}}

\end{abstract}

\section{Introduction}
\label{sec:intro}

3D head avatars have many exciting applications in immersive teleconferencing, virtual try-on, personalized video games, or education. Ideally, users can create high-quality animatable 3D head avatars from one or a few input images without expensive capture equipment or long optimization times. The avatars could even be generated from text descriptions using existing text-to-image methods.

However, creating a high-quality 3D head avatar from just a single image is extremely challenging because it is underconstrained in two regards:
(i) There are many unobserved regions complicating accurate 3D reconstruction. (ii) The model must infer realistic facial animation for a person without having seen any facial expressions of them.
These issues are typically addressed by using multi-view video recordings for training, but those are hard to obtain for sufficiently many persons. 
Many existing approaches therefore rely on monocular portrait video datasets scraped from the internet because they offer broad identity coverage and in-the-wild variability. A natural disadvantage of these datasets is that they provide only a single viewpoint per identity and typically have a strong front-view bias. As a result, models trained solely on monocular data tend to reconstruct incomplete 3D heads.\\
Despite these challenges, existing works have successfully trained single-image 3D head avatar pipelines, typically by relying heavily on geometric priors. The most common priors are 3D morphable models (3DMMs) such as FLAME~\cite{li2017flame} which provide a coarse but animatable head geometry.
In these approaches, the predicted 3D primitives, such as meshes, radiance fields, or Gaussians, are typically rigged to the 3DMM, using its deformation field to drive facial motion. This approach reduces overfitting on mono\-cular training data but limits expressiveness to the 3DMM’s predefined expression space. Still, many methods struggle with novel-view rendering.

We identify the underlying issue to be the entanglement of driving signal and target viewpoint in monocular training data. More specifically, models exploit the fact that in a monocular self-reenactment setting, the control for the facial expression is derived on the ground-truth target image itself, encouraging the model to guess the viewpoint from the expression input. Simply mixing monocular and multi-view training data does not prevent this behavior. We therefore introduce a transformer-based 3D portrait animation module with {\flextokens} that explicitly separate the model's behavior on the two dataset types. In practice, we feed learnable tokens into the transformer depending on whether a training sample stems from a monocular or a multi-view dataset. During inference, we simply use the multi-view token, prompting the model to produce a complete 3D head regardless of the input image. We further avoid relying on a restrictive 3DMM and instead learn facial expressions directly from the data, yielding more flexible animation.
Finally, to improve the quality of the renderings, we propose an upsampling architecture for the transformer based on a combination of PixelShuffle and StyleGAN~\cite{karras2020stylegan2} blocks. As a side product of our training, {\Ours} learns a smooth latent space of 3D head avatars, allowing interpolation between identities and enabling flexible fitting to arbitrary numbers of input views. Therefore, our pipeline can be used not only in a single-input scenario but also in few-shot and monocular video avatar creation settings. 

In summary, our contributions are as follows:
\begin{itemize}
    \item A novel and efficient pipeline for creating high-quality 3D head avatars from a single image
    \item Learnable {\flextokens} that combine the strengths of monocular and multi-view training to provide both strong generalization and complete 3D head avatar reconstruction
    \item An efficient upsampler architecture based on StyleGAN2 and PixelShuffle for improved visual quality
\end{itemize}

\section{Related Work}

\subsection{3D Head Avatars from Sparse Observations}

In 3D portrait animation, the goal is to predict 3D head avatars from a single image by utilizing
3D representations such as meshes~\cite{khakhulin2022rome}, Neural Radiance Fields (NeRFs)~\cite{mildenhall2021nerf, li2023hidenerf, li2023goha, chu2024gpavatar, ye2024real3dportrait, deng2024portrait4dv2, tran2024voodooxp} or 3D Gaussians (3DGS)~\cite{kerbl20233dgs, chu2024gagavatar, he2025lam, guo2025sega}, which allows rendering of novel viewpoints.
Many methods heavily rely on priors from 3D morphable models (3DMMs) such as FLAME~\cite{Blanz19993dmm, li2017flame} for coarse geometry and animation of their 3D representation. For example, both LAM~\cite{he2025lam} and GAGAvatar~\cite{chu2024gagavatar} rig 3D Gaussians to the morphable FLAME mesh, inheriting its limited animation space. In contrast, our method avoids these limitations in expressiveness by learning facial motion directly from data.

 Another line of work reconstructs avatars from one or a few observations of a person. Regression-based methods~\cite{kirschstein2025avat3r} can provide an avatar near instantly but often struggle to generalize to out-of-domain inputs or varying numbers of observations. Distillation-based methods~\cite{taubner2025cap4d, taubner2025mvp4d, xu2025vasa3d} instead use a pre-trained multi-view image or video generation network to synthesize additional views, which are then used to reconstruct a high-quality avatar~\cite{qian2024gaussianavatars}.
 While this generally improves quality, distillation is inherently slow due to the cost of invoking image or video generation. Our method generalizes well to any image domain while reconstructing high-quality avatars within minutes.
 
A different approach is to learn a photorealistic 3D head prior which can later be fitted to any set of input images ~\cite{hong2022headnerf, yang2024vrmm, yu2024one2avatar, xu2024gphm, zheng2025headgap}. These models are typically Autodecoder-based~\cite{park2019deepsdf} and trained on multi-view data. Because multi-view recordings are limited, recent methods also leverage synthetic data~\cite{saunders2025gasp, buehler2024cafca}. Our approach
also learns a latent space of avatars which can be utilized for fitting to arbitrary observations. However, we use an encoder-decoder structure, avoiding the issue of growing dictionaries in Autodecoders and enabling fast inference.

\begin{figure*}
    \centering
    \includegraphics[width=\linewidth]{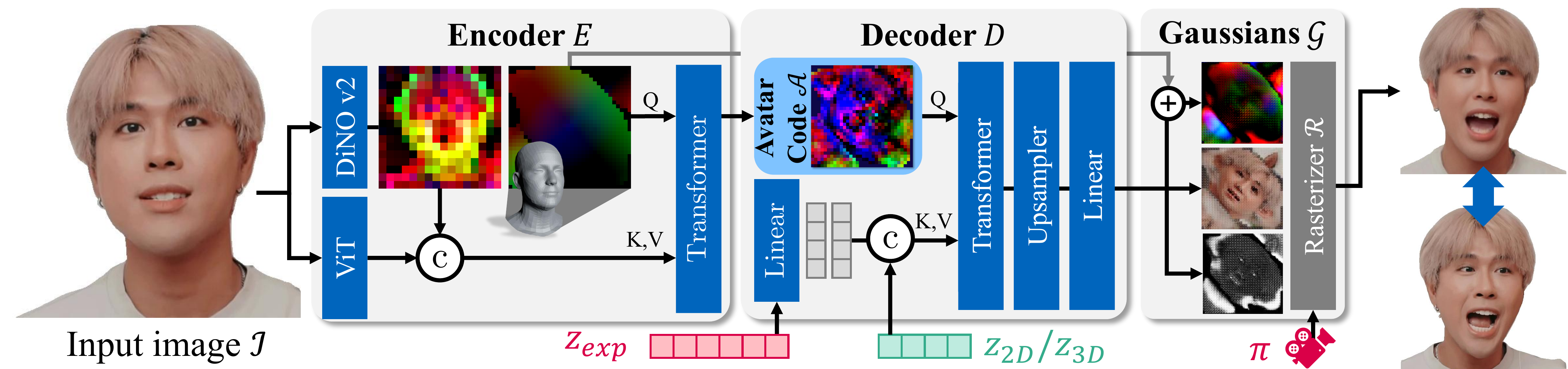}
    \caption{\textbf{Method Overview of {\Ours}}. Given the single input image $I$, our method allows to change both viewpoint $\pi$ and facial expression $z_{exp}$. The transformer-based encoder $E$ first produces a compressed avatar code $\mathcal{A}$ via cross-attention. The decoder $D$ then incorporates the effect of the facial expression $z_{exp}$ into the avatar representation. Crucially, the corresponding {\flextokens} are concatenated to the expression tokens: $z_{2D}$ if the input image $I$ comes from a monocular dataset, and $z_{3D}$ if it comes from a multi-view dataset. Finally, the upsampled avatar code is decoded into the 3D Gaussian attributes for rendering. During training, the {\flextokens} absorb data modality-specific biases such as the entanglement of driver expression and target viewpoint of monocular datasets. At inference time, only $z_{3D}$ is used to inherit the disentangled behavior of multi-view datasets yielding both generalized and complete 3D head avatars. }
    \label{fig:3_method_overview}
\end{figure*}

\subsection{Learnable Dataset Embeddings}

Learnable embeddings or tokens are used in several settings.
In task adaptation, they are used for parameter-efficient finetuning~\cite{lester2021softtokens, sandler2022finetunetaskadaptation, jia2022visualprompttuning}. In multi-dataset training, dataset indicators help unify heterogeneous datasets into a shared feature space~\cite{meng2023detectionhub, zheng2022promptvisiontransformer}. Multi-modal transformers similarly use modality-specific embeddings to distinguish input types~\cite{jaegle2021perceiver, zhu2022uniperceiver}.
In 3D reconstruction, NeRF-in-the-wild~\cite{martin2021nerfinthewild} learns a per-image embedding that captures aspects of the input that the subsequent generalized NeRF cannot explain. Similarly, methods like Nerfies~\cite{park2021nerfies} or Cafca~\cite{buehler2024cafca} bake unwanted temporal variations of the input images into learnable embeddings. 

The difference in our setting is that we introduce dataset-level embeddings to explicitly capture dataset-induced biases. This allows us to suppress these biases at inference time, enabling a model trained on mixed monocular and multi-view data to behave as if it were supervised by multi-view observations alone while keeping the generalization capabilities induced by the monocular training data.

\section{Method}

Given a single portrait image $I$, our goal is to create an animatable avatar representation $\mathcal{A}$ which we can control via animation codes $z_{exp}$ and render from arbitrary viewpoints. A visual overview of our approach is depicted in~\cref{fig:3_method_overview}. We adopt an encoder-decoder perspective and split the image synthesis process into multiple stages: (1) An Encoder $E$ that finds a suitable avatar code $\mathcal{A}$ based on the input image $I$, (2) a decoder $D$ that creates a set of articulated 3D Gaussians given an expression code $z_{exp}$, and (3) a renderer $\mathcal{R}$ which renders the 3D Gaussian representation from the desired viewpoint $\pi$:
\begin{align}
    \mathcal{A} &= E(I) \\
    \mathcal{G} &= D(\mathcal{A}, z_{exp}) \\
    I^{pred} &= \mathcal{R}(\mathcal{G}, \pi)
\end{align}
In practice, we use the tile-based differentiable rasterizer from 3DGS~\cite{kerbl20233dgs} as $\mathcal{R}$ and expression codes $z_{exp}$ from FLAME~\cite{li2017flame}. $E$ and $D$ are implemented via transformers~\cite{vaswani2017attention}, and $\mathcal{A} \in \mathbb{R}^{H_l \times W_l \times D}$ is a 2-dimensional latent code that lives in the UV-space of a template head mesh. 

Crucially, the encoder-decoder design choice leads to the emergence of a smooth latent space of avatars during training. This enables applications that go beyond direct feed-forward prediction of a 3D head avatar from a single image (see~\cref{sec:avatar_latent_space}).

\subsection{Encoder $E$: Projecting onto an Avatar manifold}
The general design of our encoder is inspired by LAM~\cite{he2025lam} with the focus on producing a compressed avatar representation. 
For this purpose, we employ a head template mesh with corresponding UV space which will host the avatar code's features. 
We begin by first extracting image features $f_{img}$ with a pre-trained DINOv2~\cite{oquab2023dinov2} model and a shallow learnable ViT.
\begin{align}
    f_{img} &= \textsc{MLP}([\textsc{DINO}(I), \textsc{ViT} ([I,  I^{pluck}])])
\end{align}
where $I^{pluck}$ are the plucker embeddings of the camera viewpoint of the input image $I$.
To map the image features into the template's UV space, we define queries $Q$ anchored in UV space. This is done by uniformly sampling 3D surface positions in the UV space of the template mesh $\mathcal{T}$ and encoding them with sinusoidal frequencies:
\begin{align}
    x_{mesh}, x_{uv} &\leftarrow \mathcal{T} \\
    Q &= \textsc{pe}(x_{mesh})
\end{align}
Finally, we perform cross-attention from the UV-anchored queries $Q$ to the image features $f_{img}$:
\begin{align}
    \mathcal{A} &= \textsc{Attention}(Q, f_{img}, f_{img})
\end{align}
In practice, we use the attention implementation from MMDIT~\cite{esser2024mmdit}. The result is a compact 2-dimensional latent code $\mathcal{A} \in \mathbb{R}^{H_l \times W_l \times D}$ that contains all relevant information from the input image but is agnostic to both viewpoint and facial expression.

\begin{figure}[t]
    \centering
    \includegraphics[width=\linewidth]{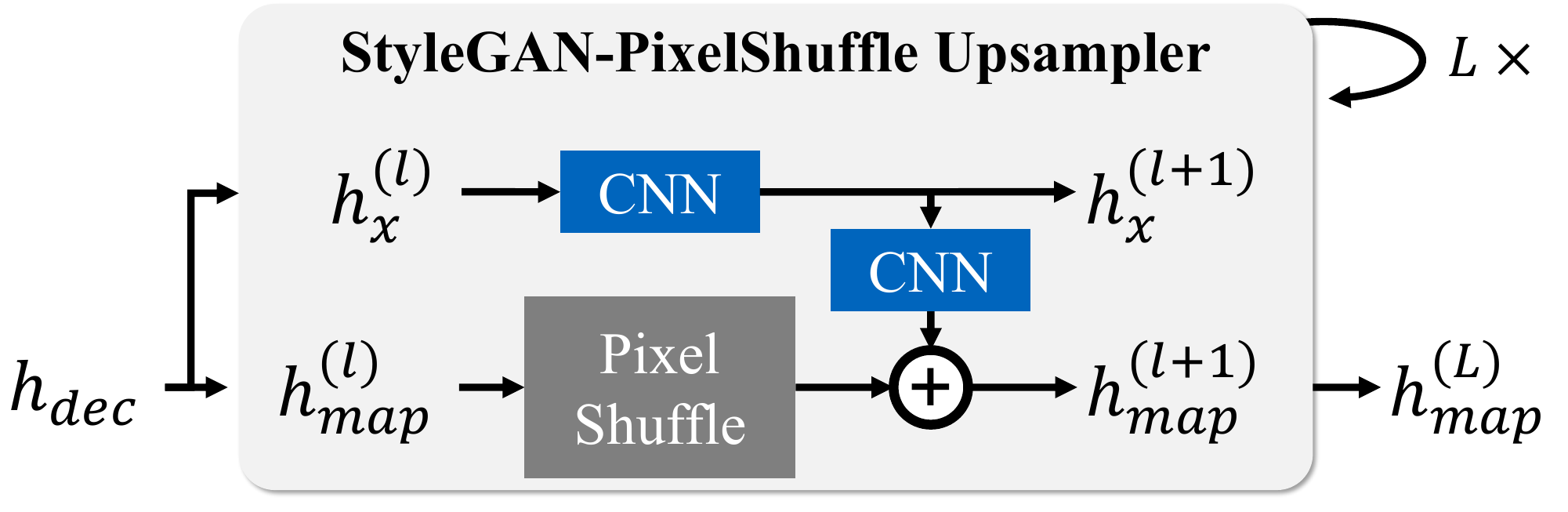}
    \caption{\textbf{Architecture of the StyleGAN-PixelShuffle block.}}
    \label{fig:3_stylegan_pixelshuffle}
\end{figure}

\subsection{Decoder $D$: Decode Articulated 3D Gaussians}
The decoder's goal is to incorporate the effect of facial expressions on the avatar representation and to produce the final 3D Gaussians for rendering. For animation modeling, we adopt the approach of Avat3r~\cite{kirschstein2025avat3r} and use cross-attention from the internal representation to a sequenced expression code $s_{exp} \in \mathbb{R}^{N_{exp} \times D}$. This model-free approach allows the network to learn facial animations from the data without being limited to the animation space of a pre-defined 3D face model:
\begin{align}
    s_{exp} &= \textsc{MLP}(z_{exp}) \\
    h_{dec} &= \textsc{Attention}(\mathcal{A}, s_{exp}, s_{exp})
\end{align}
The expression code $z_{exp}$ can be any description of the facial state, such as audio, 3DMM coefficients, or an image embedding derived from a driving image. In practice, we use the expression codes of FLAME~\cite{li2017flame}. However, note that our network design makes no assumptions about the structure of $z_{exp}$ and can easily be applied to different driving signals.

The resulting decoder feature map $h_{dec} \in \mathbb{R}^{H_l \times W_l \times D}$ is then upsampled $L$ times yielding $h_{map}^{(L)} \in \mathbb{R}^{L\cdot H_l \times L \cdot W_l \times \frac{D}{L^2}}$. This is crucial for decoding sufficiently many 3D Gaussians. \cref{fig:3_stylegan_pixelshuffle} shows an overview of our upsampler design which uses a combination of PixelShuffle~\cite{shi2016pixelshuffle} and CNN blocks inspired by StyleGAN2~\cite{karras2020stylegan2}:
\begin{align}
    h_{x}^{(l + 1)} &= \textsc{Cnn}\left(h_{x}^{(l)}\right) \\
    h_{map}^{(l + 1)} &= \textsc{PixelShuffle}\left(h_{map}^{(l)}\right) + \textsc{Cnn}\left(h_{x}^{(l + 1)}\right)
\end{align}
with $h_x^{(0)} = h_{map}^{(0)} = h_{dec}$. This is followed by bilinear grid sampling to extract one feature per Gaussian:
\begin{align}
    x &= \textsc{GridSample}\left(h_{map}^{(L)}, x_{uv}\right)
\end{align}
where $x_{uv}$ are the texel locations of the sampled points $x_{mesh}$ on the template mesh.
In practice, we use $L = 2$ upsampling steps and perform grid sampling with another 2x upsampling, yielding a total upsampling rate of 8x. The resulting features $x \in \mathbb{R}^{G \times \frac{D}{L^2}}$ hold information for each 3D Gaussian that are decoded with an MLP:
\begin{align}
    \mathcal{G} &= \textsc{MLP}(x)
\end{align}
We also initialize the Gaussians' positions on the template mesh surface $x_{mesh}$:
\begin{align}
    \mathcal{G}_{pos} &\leftarrow \mathcal{G}_{pos} + x_{mesh}
\end{align}
The final 3D Gaussians $\mathcal{G}$ can then be rendered via the tile-based rasterizer of~\cite{kerbl20233dgs}:
\begin{align}
    I_{pred} = \mathcal{R}(\mathcal{G}, \pi)
\end{align}
In practice, we use the batched rendering implementation of gsplat~\cite{ye2025gsplat} for better training performance.

\begin{figure}
    \centering
    \includegraphics[width=\linewidth,trim={0 0.2cm 0 0},clip]{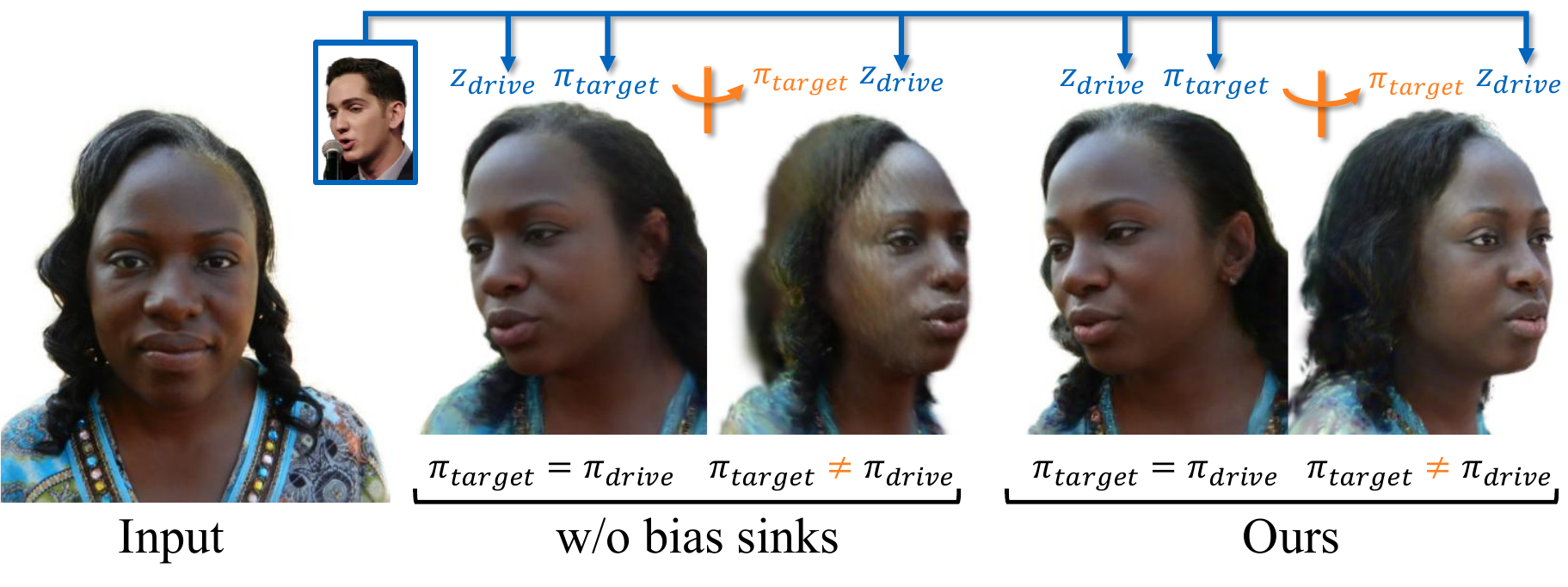}
    \caption{\textbf{Entanglement of driving signal and target viewpoint.} Naive training on monocular data works well as long as both expression code $z_{drive}$ and rendering camera $\pi_{target}$ are transferred to the avatar ($\pi_{target} = \pi_{drive}$). Artifacts occur when the rendering camera is moved, i.e., rendering and driving viewpoint differ ($\pi_{target} \neq \pi_{drive}$). This issue is fixed by our proposed {\flextokens}.}
    \label{fig:3_entanglement}
\end{figure}

\subsection{Fighting Entanglement with Bias Sinks}
During training, 3D portrait animation models minimize an image loss $\mathcal{L}\left(f(I_{source}, z_{target}), I_{target}\right)$  where the expression code $z_{target} = \textsc{Track}(I_{drive})$ is derived from a driving image that matches the target expression.
In monocular video datasets, $I_{drive} = I_{target}$ since there is only a single camera available. In this case, the derived expression code $z_{target}$ can leak information about the viewpoint $\pi_{target}$ of the target image.
The model may exploit this by predicting only a partial 3D head which is sufficient to satisfy the loss from that specific viewpoint.
We refer to this failure mode as \textit{entanglement of driving signal and target viewpoint}. Although acceptable when $\pi_{target} = \pi_{drive}$ (e.g., standard portrait animation), it breaks in applications requiring free-view rendering ($\pi_{target} \neq \pi_{drive}$), leading to incomplete heads as seen in~\cref{fig:3_entanglement}.

Multi-view datasets break this entanglement by providing multiple viewpoints for the same facial expression, but they are too limited in scale for good generalization. To address this, we introduce {\flextokens}, which are two learnable tokens $z_{2D}$ and $z_{3D}$ that are concatenated to the expression code sequence $s_{exp}$ before decoding:
\begin{align}
    s_{exp} \leftarrow [s_{exp}, z_{bias}]
\end{align}
During training, samples from monocular datasets use $z_{2D}$ and multi-view samples use $z_{3D}$.
This makes the decoder explicitly aware of a sample's provenance absorbing the bias of a particular dataset type. In practice, the model learns to predict incomplete 3D heads whenever it sees the $z_{2D}$ token and produces a complete avatar when $z_{3D}$ is given. Crucially, this design still allows the model to share knowledge across dataset types. In particular, when feeding in $z_{3D}$, the model still benefits from the generalization obtained from the monocular video training. During inference, we always feed in the $z_{3D}$ token to obtain both well generalized and complete 3D head avatars from a single image.

\newcommand{\dsetcvt}{\small CV-T}
\newcommand{\dseth}{\small H3}
\newcommand{\dsetcafca}{\small C}
\newcommand{\dsetnersemble}{\small N}
\newcommand{\dsetava}{\small A}

\begin{table*}[t]
    \centering
    \resizebox{\linewidth}{!}{
    \begin{tabular}{lrrrllrl}
        \toprule
        Task & \#Inputs & Output assumption & \multicolumn{2}{c}{Training data} & Evaluation data & Fitting & Figures\\
        \midrule
        3D Portrait Animation (§\ref{sec:3d_portrait_animation})
            & 1 & $\pi_{target} = \pi_{drive}$ & \multirow{4}{*}{\begin{minipage}{0.13\linewidth}\centering $ \left. \begin{aligned}  \textrm{CelebV-Text} \\[-0.5em] \textrm{Hallo3} \\[-0.5em] \textrm{NeRsemble} \\[-0.5em] \textrm{Cafca}  \end{aligned} \right\} $\end{minipage} + } 
            & Ava256 & VFHQ-Test & 200 steps & \cref{tab:4_vfhq_comparison}  \\
        Single-image Avatar Creation (§\ref{sec:single_image_avatar_creation})
            & 1 & $\pi_{target} \neq \pi_{drive}$ & 
            & - & Ava256\textsuperscript{5 persons} & 200 steps & \cref{tab:main_comparison}, \cref{fig:4_main_comparison} \\
        Few-shot Avatar Creation (§\ref{sec:few_shot_avatar_creation})
            & 4 & $\pi_{target} \neq \pi_{drive}$ & 
            & Ava256\textsuperscript{Avat3r train} & Ava256\textsuperscript{Avat3r test} & 1000 steps & \cref{tab:4_avat3r_comparison} \\
        Monocular Avatar Creation (§\ref{sec:monocular_avatar_creation})
            & 900 & $\pi_{target} \neq \pi_{drive}$ & 
            & Ava256 & NeRSemble Benchmark & 2000 steps & \cref{tab:4_nersemble_benchmark_comparison}, \cref{fig:4_nersemble_benchmark} \\
        \bottomrule
    \end{tabular}
    }
    \caption{\textbf{Overview of Experimental Results.} We evaluate {\Ours} on 4 different tasks and 3 different datasets. }
    \label{tab:experiment_overview}
\end{table*}

\subsection{Training with Perceptual Losses}
We use the L1 and SSIM losses from 3DGS:
\begin{align}
    \mathcal{L}_1 &= \Vert I_{pred} - I_{target} \Vert_1 \\
    \mathcal{L}_{SSIM} &= 1 - \textsc{SSIM}(I_{pred}, I_{target})
\end{align}
Inspired by PercHead~\cite{oroz2025perchead}, we additionally employ perceptual losses based on DINOv2~\cite{oquab2023dinov2} and the Segment Anything Model (SAM)~\cite{ravi2024sam2}:
\begin{align}
    \mathcal{L}_{DINO} &= \Vert \textsc{DINO}_f(I_{pred}) - \textsc{DINO}_f(I_{target}) \Vert_1 \\
    \mathcal{L}_{SAM} &= \Vert \textsc{SAM}_f(I_{pred}) - \textsc{SAM}_f(I_{target}) \Vert_1
\end{align}
where $\textsc{DINO}_f(.)$ and $\textsc{SAM}_f(.)$ extract intermediate feature maps of the given image.
The final reconstruction loss is a combination of all terms:
\begin{align}
    \mathcal{L}_{rec} = \mathcal{L}_1 + \mathcal{L}_{SSIM} + \mathcal{L}_{DINO} + \mathcal{L}_{SAM}
\end{align}

\subsection{Fitting $A$ to Additional Observations}
\label{sec:avatar_latent_space}
Often, more than one image of a person is available, e.g., a set of images $(\mathcal{I}^{many}, z^{many}_{exp}, \pi^{many})$ with corresponding expression codes and cameras.
We can use our encoder $E$ to get an initialization for $\mathcal{A}$ by using any one of the available observations:
\begin{align}
    \mathcal{A}^{init} &= E\left(\mathcal{I}^{many}_0\right)
\end{align}
This initial estimate of the avatar can then be optimized by fitting it against all observations:
\begin{align}
    \mathcal{I}^{many}_{pred} &= \mathcal{R}(D(\mathcal{A}^{init}, z^{many}_{exp}), \pi^{many}) \\
    \mathcal{L}_{fit} &= \mathcal{L}_{rec}(\mathcal{I}^{many}_{pred}, \mathcal{I}^{many})
\end{align}
By minimizing $\mathcal{L}_{fit}$, one can obtain an animatable 3D head avatar representation $\mathcal{A}^{fit}$ that incorporates all available observations of the person. Crucially, we only make $\mathcal{A}^{init}$ learnable and keep the entire decoder $\mathcal{D}$ fixed to avoid overfitting on the sparse inputs.

This procedure is similar to how autodecoder-style photorealistic 3D head models such as GPHM~\cite{xu2024gphm}, HeadGAP~\cite{zheng2025headgap}, or HeadNeRF~\cite{hong2022headnerf} create an avatar of an person. However, our approach has two advantages: First, it can be trained on mostly monocular video datasets whereas autodecoder-style models typically require multi-view training. Second, our approach also has an encoder which speeds up the optimization process by providing already an initial guess of the latent avatar code.

\section{Experimental Results}

\subsection{Training}

We train {\Ours} on 5 datasets: 2 monocular portrait video datasets (CelebV-Text~\cite{yu2023celebvtext} and Hello3~\cite{cui2025hallo3}), 2 multi-view datasets (NeRSemble~\cite{kirschstein2023nersemble} and Ava256~\cite{martinez2024ava256}), and the synthetic multi-view Cafca dataset~\cite{buehler2024cafca}. We sample 40k clips from the monocular sets, use all Ava256 recordings, $\sim$25\% of NeRSemble, and neutral-expression frames from all Cafca identities. While monocular data provides generalization, NeRSemble and Ava256 offer high-quality expressions, and Cafca supplies full 360° supervision.

We extract cameras $\pi$ and expression codes $z_{exp}$ using Pixel3DMM~\cite{giebenhain2025pixel3dmm}. For NeRSemble and Ava256, we only track the frontal camera.
Training uses Adam~\cite{kingma2014adam} with a learning rate of $1e-4$. Perceptual losses are introduced after 400k steps to avoid early overfitting to high-frequency details. In total, the model is trained for 1M steps with a batch size of 20 on one A100 GPU, taking roughly 3 weeks.

\begin{table*}[t]
\centering

\resizebox{\textwidth}{!}{
\begin{tabular}{lrrrrrrrrrr}
\toprule
 & \multicolumn{7}{c}{\textbf{Self Reenactment}} & \multicolumn{3}{c}{\textbf{Cross Reenactment}} \\
\cmidrule(lr){2-8} \cmidrule(lr){9-11}
 & PSNR$\uparrow$ & SSIM$\uparrow$ & LPIPS$\downarrow$ & CSIM$\uparrow$ & AED$\downarrow$ & APD$\downarrow$ & AKD$\downarrow$ & CSIM$\uparrow$ & AED$\downarrow$ & APD$\downarrow$ \\
\midrule
    GPAvatar~\cite{chu2024gpavatar}      
        & 21.04 & 0.807 & 0.150 & 0.772 & 0.132 & 0.189 & 4.226 & 0.564 & 0.255 & 0.328 \\
    Real3DPortrait~\cite{ye2024real3dportrait} 
        & 20.88 & 0.780 & 0.154 & 0.801 & 0.150 & 0.268 & 5.971 & \textbf{0.663} & 0.296 & 0.411 \\
    Portrait4D~\cite{deng2024portrait4d}  
        & 20.35 & 0.741 & 0.191 & 0.765 & 0.144 & 0.205 & 4.854 & 0.596 & 0.286 & 0.258 \\
    Portrait4D-v2~\cite{deng2024portrait4dv2} 
        & 21.34 & 0.791 & 0.144 & 0.803 & 0.117 & 0.187 & 3.749 & 0.656 & 0.268 & 0.273 \\
    GAGAvatar~\cite{chu2024gagavatar}       
        & 21.83 & 0.818 & 0.122 & 0.816 & 0.111 & 0.135 & 3.349 & 0.633 & 0.253 & 0.247 \\
    LAM~\cite{he2025lam}                     
        & 22.65 & 0.829 & 0.109 & 0.822 & 0.102 & 0.134 & \textbf{2.059} & 0.651 & 0.250 & 0.356 \\
    Ours %
        & \textbf{23.47} & \textbf{0.837} & \textbf{0.099} & \textbf{0.830} & \textbf{0.075} & \textbf{0.010} & 2.965 & \textbf{0.663} & \textbf{0.223} & \textbf{0.026} \\
\bottomrule
\end{tabular}
}
\caption{\textbf{3D Portrait Animation comparison on the VFHQ dataset.} We evaluate the ability to animate a single image by transferring facial motion and head pose from a driving video showing the same person (self-reenactment) or a different person (cross-reenactment). }
\label{tab:4_vfhq_comparison}
\end{table*}

\subsection{Experiment Setup}

\paragraph{Tasks.} 
\cref{tab:experiment_overview} shows an overview of our experiment setup.
We evaluate {\Ours}'s ability to create 3D head avatars in a variety of situations: 

\textit{3D Portrait Animation.} 
In this well-established task, the goal is to animate a portrait image by transferring both expression and head pose from a second image (which can be of a different person). In this setting, methods can exploit the entanglement of driving signal and target viewpoint since $\pi_{target} = \pi_{drive}$.

\textit{Single-image 3D Head Avatar Creation.}
Similar to 3D Portrait animation, a single image is given with the additional requirement to be able to freely change the camera viewpoint. In this setting, no connection between the driving signal and the rendering viewpoint can be assumed since $\pi_{target} \neq \pi_{drive}$. 

\textit{Few-shot 3D Head Avatar Creation.}
In this task, 4 images of a person are provided with the goal to create a complete 3D head avatar that can be freely animated and rendered from any viewpoint.

\textit{Monocular 3D Head Avatar Creation.}
For the last task, one or several monocular videos of a person are available to create a 3D head avatar. 
We compare against recent state-of-the-art methods on the public leaderboard of the NeRSemble benchmark.

\paragraph{Metrics.} 
Across all our experiments, we employ three paired-image metrics to measure the quality of individual rendered images: Peak Signal-to-Noise Ratio (PSNR), Structural Similarity Index (SSIM)~\cite{wang2004ssim}, and Learned Perceptual Image Patch Similarity (LPIPS)~\cite{zhang2018lpips}. 
Furthermore, we make use of two face-specific metrics: Average Keypoint Distance (AKD) measured in pixels with keypoints estimated from PIPNet~\cite{jin2021pipnet}, and cosine similarity (CSIM) of identity embeddings based on ArcFace~\cite{deng2019arcface}.
Temporal consistency is measured with FovVideoVDP~\cite{mantiuk2021fovvideovdp} (JOD) which is sensitive to flickering, noise and other temporal artifacts.
Finally, we estimate 3DMM coefficients using the forward regressor of~\cite{deng2019deep3dfacerecon} to compute Average Expression Distance (AED) and Average Pose Distance (APD) by computing the L1 distance of the corresponding 3DMM coefficients.

\subsection{3D Portrait Animation}
\label{sec:3d_portrait_animation}
We follow the evaluation protocol of GAGAvatar~\cite{chu2024gagavatar} and evaluate both self-reenactment and cross-reenactment performance on the VFHQ test split~\cite{xie2022vfhq}. The results can be seen in~\cref{tab:4_vfhq_comparison}. Our method improves in all metrics except AKD over the previous state-of-the-art. This shows that {\Ours} can generalize well to unseen persons and can animate portraits with different driving persons. 

\begin{figure*}
    \centering
    \includegraphics[width=\linewidth]{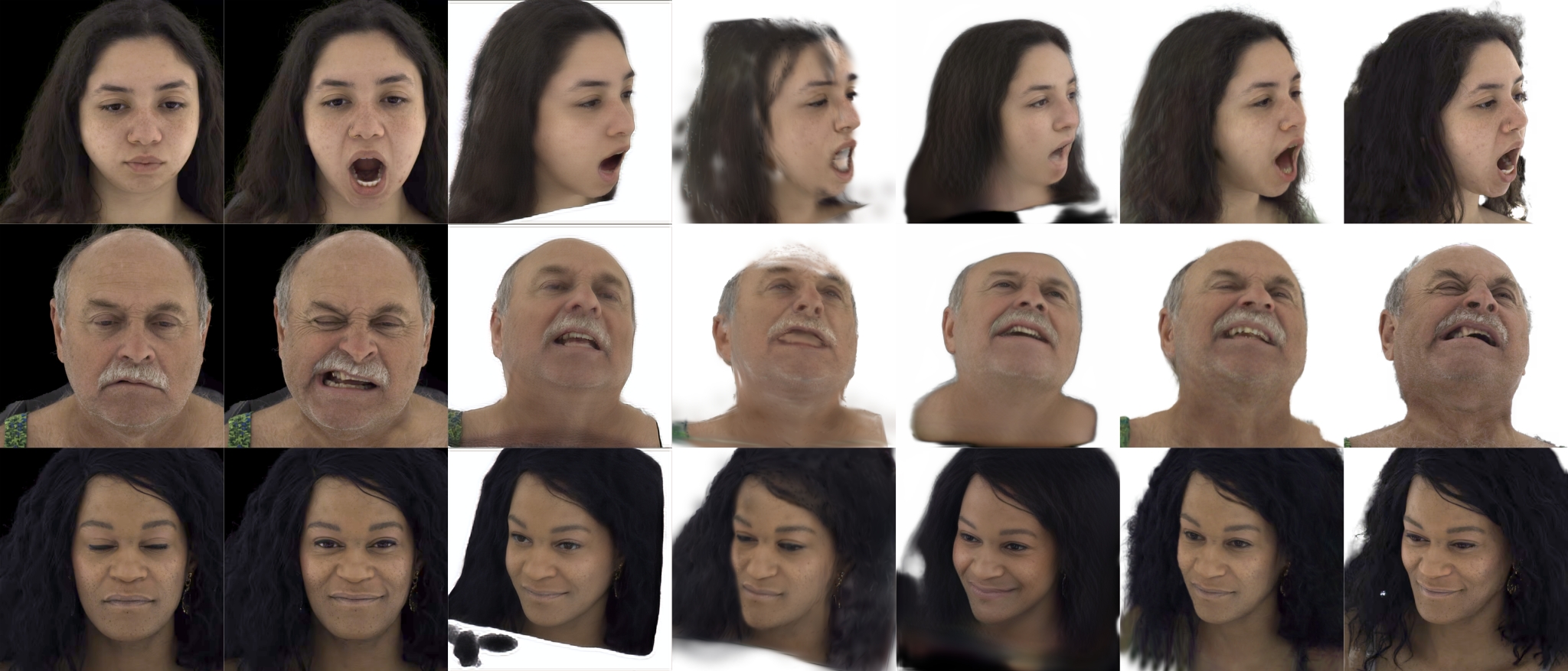}
    \begin{tabularx}{\linewidth}{YYYYYYY}
        Input & Driver & Portrait4Dv2~\cite{deng2024portrait4dv2} & LAM~\cite{he2025lam} & GAGAvatar~\cite{chu2024gagavatar} & Ours & GT \\
    \end{tabularx}
    \caption{\textbf{Qualitative Single-image Avatar Creation comparison on the Ava256 dataset.} We compare our method to the recent state-of-the-art on 3D head avatar creation from a single portrait image. Our method produces more complete 3D head avatars and re-enacts the target expression more faithfully.}
    \label{fig:4_main_comparison}
\end{figure*}

\begin{table}[tb]
    \centering
    \addtolength{\tabcolsep}{-0.2em}
    \resizebox{\linewidth}{!}{
    \begin{tabular}{llrrrrr}
        \toprule
         &&  \small PSNR$\uparrow$ & \small SSIM$\uparrow$ & \small LPIPS$\downarrow$ & \small AKD$\downarrow$ & \small CSIM$\uparrow$ \\
        \midrule
        \multirow{4}{*}{\rotatebox[origin=c]{90}{\parbox[c]{1.7cm}{\centering \small Single-image}}}
        &Portrait4Dv2~\cite{deng2024portrait4dv2} 
            & 11.9 &  0.671 & 0.404 & 7.77 & 0.578 \\
        & LAM~\cite{he2025lam} %
            & 13.1 & 0.702 & 0.399 & 11.2 & 0.411 \\
        & GAGAvatar~\cite{chu2024gagavatar}
            & 12.7 & 0.709 & 0.371 & 7.45 & 0.555 \\
        & Ours %
            & \textbf{16.9} & \textbf{0.762} & \textbf{0.265} & \textbf{5.52} & \textbf{0.695} \\
        \midrule
        \multirow{4}{*}{\rotatebox[origin=c]{90}{\parbox[c]{1.5cm}{\centering \small Few-shot}}}
        & InvertAvatar~\cite{zhao2024invertavatar}
            & 13.0 & 0.288 & 0.590 & 52.3 & 0.296 \\
        & GPAvatar~\cite{chu2024gpavatar}
            & 20.0 & 0.700 & 0.291 & 5.72 & 0.341 \\
        & Avat3r~\cite{kirschstein2025avat3r}
            & 20.8 & 0.715 & 0.310 & 5.66 & 0.616 \\
        & Ours %
            & \textbf{21.1} & \textbf{0.733} & \textbf{0.218} & \textbf{5.39} & \textbf{0.755} \\
        \bottomrule
    \end{tabular}
    }
    \caption{\textbf{Single-image and Few-shot Avatar Creation comparison on the Ava256 dataset.}}
    \label{tab:main_comparison}
    \label{tab:4_avat3r_comparison}
\end{table}

\subsection{Single-image 3D Head Avatar Creation}
\label{sec:single_image_avatar_creation}

We evaluate single-image 3D head avatar reconstruction on the Ava256 dataset~\cite{martinez2024ava256}. We select one challenging sequence for 5 diverse subjects. The frontal frame of the first timestep serves as input, and we uniformly sample 10 target expressions from 4 distinct cameras per sequence, yielding 200 test images. Expression codes $z_{exp}$ are extracted from the frontal view, unlike standard 3D portrait animation settings where the driving and target viewpoints coincide. This setup is more demanding as methods that exploit viewpoint information in $z_{exp}$ are penalized. This evaluation better reflects real applications that require freely animating an avatar without assumptions about the rendering viewpoint.

Results in~\cref{tab:main_comparison} and~\cref{fig:4_main_comparison} show that our method substantially outperforms recent approaches, producing realistic, complete, and expressive 3D heads. For fairness, the entire Ava256 dataset is held out during training. 
Note that the publicly released version of LAM used in the comparison was trained on both monocular (VFHQ~\cite{xie2022vfhq}) and multi-view (NeRSemble~\cite{kirschstein2023nersemble}) data. Hence, our gains cannot be attributed solely to multi-view supervision. Further analysis is provided in the ablation section.

\subsection{Few-shot 3D Head Avatar Creation}
\label{sec:few_shot_avatar_creation}
Thanks to {\Ours}'s smooth avatar latent space, we can seamlessly integrate multiple observations of a subject via fitting following~\cref{sec:avatar_latent_space}. We evaluate this on the Ava256 dataset using the same protocol as Avat3r~\cite{kirschstein2025avat3r}: 4 input images of a person are provided, and the model must render a novel expression from a novel viewpoint. 
To build an avatar, we encode one of the 4 images to obtain an initial code $\mathcal{A}^{init}$ and then optimize it for 1000 steps ($\sim$7 minutes per avatar) to match all four inputs. For fair comparison, we train with Ava256 but exclude all test identities, following Avat3r. Metrics are computed on a subset of sequences where Pixel3DMM tracking succeeds. 
As shown in~\cref{tab:4_avat3r_comparison}, our method outperforms Avat3r, particularly in sharpness (LPIPS) and identity preservation (CSIM).

\begin{table}[t]
\centering
\addtolength{\tabcolsep}{-0.2em}
\resizebox{\linewidth}{!}{
\begin{tabular}{lrrrrrr}
\toprule
 & \footnotesize{PSNR}$\uparrow$ & \footnotesize{SSIM}$\uparrow$ & \footnotesize{LPIPS}$\downarrow$ & \footnotesize{JOD}$\uparrow$ & \footnotesize{AKD}$\downarrow$ & \footnotesize{CSIM}$\uparrow$ \\
\midrule
    INSTA~\cite{zielonka2023insta}
        & 15.8 & 0.771 & 0.344 & 4.83 & 5.22 & 0.631 \\
    FlashAvatar~\cite{xiang2024flashavatar} 
        & 16.3 & 0.731 & 0.386 & 4.15 & 19.18 & 0.304 \\
    TaoAvatar~\cite{chen2025taoavatar}
        & 18.2 & 0.789 & 0.267 & 5.28 & 5.50 & 0.715 \\
    FATE~\cite{zhang2025fate}
        & 19.1 & 0.820 & 0.220 & 5.56 & 3.52 & 0.770 \\
    HRAvatar~\cite{zhang2025hravatar} 
        & 19.5 & 0.817 & 0.214 & 5.76 & 4.62 & 0.765 \\
    CAP4D~\cite{taubner2025cap4d}
        & 19.8 & 0.821 & 0.185 & 5.79 & 4.19 & 0.793 \\
    RGBAvatar~\cite{li2025rgbavatar}
        & 20.6 & 0.829 & 0.181 & 6.03 & \textbf{3.41} & 0.824 \\
    Ours %
        & \textbf{20.9} & \textbf{0.830} & \textbf{0.156} & \textbf{6.08} & 3.80 & \textbf{0.827} \\
\bottomrule
\end{tabular}
}
\caption{\textbf{Monocular Avatar Creation comparison on the NeRSemble Benchmark.} We evaluate the ability to render novel views and novel expressions given monocular videos of 5 persons.}
\label{tab:4_nersemble_benchmark_comparison}
\end{table}

\subsection{Monocular 3D Head Avatar Creation}
\label{sec:monocular_avatar_creation}
\begin{figure}[t]
    \centering
    \includegraphics[width=\linewidth]{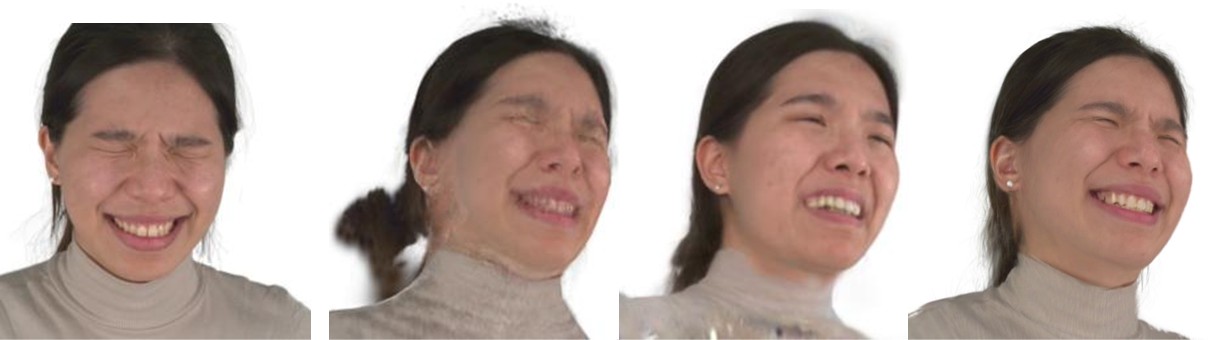}
    \begin{tabularx}{\linewidth}{YYYY}
        Driver & RGBAvatar & CAP4D & Ours
    \end{tabularx}
    \caption{\textbf{Comparison on the NeRSemble Benchmark.}}
    \label{fig:4_nersemble_benchmark}
\end{figure}

Finally, we evaluate the scalability of {\Ours} on the NeRSemble monocular 3D head avatar benchmark~\cite{kirschstein2023nersemble}, which requires creating 3D avatars from video clips of 5 subjects. As in our few-shot experiments, we predict an initial avatar code $\mathcal{A}^{init}$ and fit it to 900 evenly sampled frames from the training videos for 2000 iterations ($\sim$10 minutes per avatar). No benchmark subject data is used during training. Results in~\cref{tab:4_nersemble_benchmark_comparison} show that we outperform all baselines on nearly all metrics, with significant gains in sharpness (LPIPS). A visual comparison is shown in~\cref{fig:4_nersemble_benchmark}. Notably, our method surpasses CAP4D~\cite{taubner2025cap4d}, that relies on a strong multi-view 3D head prior, while using fewer frames and achieving much faster fitting (10 minutes vs. 4 hours).

\begin{figure*}
    \centering
     \includegraphics[width=\linewidth]{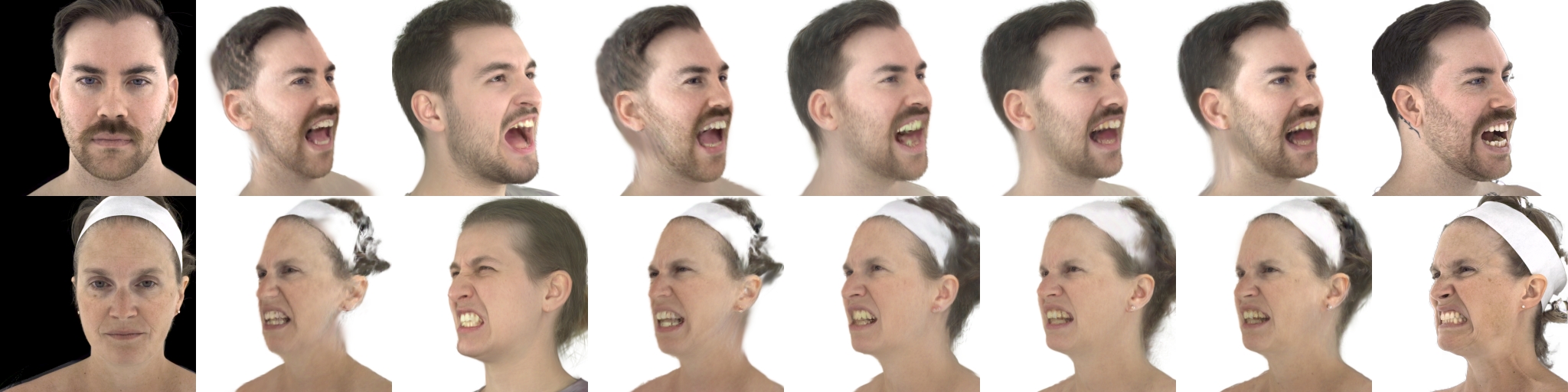}
    \begin{tabularx}{\linewidth}{YYYYYYYYY}
        \small{Input} & \small{Only 2D} & \small{Only 3D} & \small{w/o~bias~sinks} & \small{\mbox{w/o~StyleGAN}} & \small{Ours\textsuperscript{ref}} & \small{Ours + fitting} & \small{GT} \\
    \end{tabularx}
    \caption{\textbf{Qualitative Ablation of method components on the Ava256 dataset.}}
    \label{fig:4_ablations}
\end{figure*}

\subsection{Ablations}
In~\cref{tab:4_ablation} and~\cref{fig:4_ablations}, we present ablations of our dataset and architecture choices.  
The ablations are compared on the Ava256 dataset on the single-image 3D head avatar creation task. Crucially, we hold out the entire Ava256 dataset from training to measure performance on an unseen data domain.

\newcommand{\checkbox}{\scalebox{1.5}{$\boxtimes$}}
\newcommand{\emptybox}{\scalebox{1.5}{$\square$}}

\begin{table}[t]
    \centering
    \addtolength{\tabcolsep}{-0.4em}
    \resizebox{\linewidth}{!}{
    \begin{tabular}{lcccccrrrrr}
    \toprule
      &2D&3D&$\mathcal{B}$&$\mathcal{U}$&$\mathcal{F}$&  \small PSNR$\uparrow$ & \small SSIM$\uparrow$ & \small LPIPS$\downarrow$ & \small AKD$\downarrow$ & \small CSIM$\uparrow$ \\
    \midrule
        only 2D
            & \checkbox & \emptybox & \emptybox & \checkbox & \emptybox
            & 13.7 & 0.736 & 0.358 & 6.59 & 0.593 \\
        only 3D
            & \emptybox & \checkbox & \emptybox & \checkbox & \emptybox
            & 13.2 & 0.699 & 0.378 & 10.4 & 0.119 \\
        w/o {\flextokens}
            & \checkbox & \checkbox & \emptybox & \checkbox & \emptybox
            & 14.5 & 0.747 & 0.351 & 5.98 & 0.583 \\
        w/o StyleGAN
            & \checkbox & \checkbox & \checkbox & \emptybox & \emptybox
            & 17.1 & 0.765 & 0.287 & 7.03 & 0.614 \\
        Ours\textsuperscript{ref}
            & \checkbox & \checkbox & \checkbox & \checkbox & \emptybox
            & \textbf{17.2} & 0.768 & 0.285 & 6.34 & 0.621 \\
        Ours + fitting
            & \checkbox & \checkbox & \checkbox & \checkbox & \checkbox
            & 16.9 & \textbf{0.771} & \textbf{0.280} & \textbf{5.59} & \textbf{0.682} \\
    \bottomrule
    $\mathcal{B} = $ bias sinks & \multicolumn{8}{r}{$\mathcal{U} =$ StyleGAN-PixelShuffle Upsampler} & \multicolumn{2}{r}{$\mathcal{F}=$ Fitting}
    \end{tabular}
    }
    \caption{\textbf{Quantitative Ablation on the Ava256 dataset.} Ablation models are only trained for 500k iterations to save compute resources. Hence, the numbers for \textit{Ours + fitting} differ slightly from \textit{Ours} in~\cref{tab:main_comparison} even though both use the same evaluation setup.}
    \label{tab:4_ablation}
\end{table}

\paragraph{Effect of Training Data.}
Training only on monocular data (\textit{only 2D}) produces partial 3D heads due to entanglement between driving signal and target viewpoint. Multi-view training (\textit{only 3D}) resolves this, yielding complete high-quality avatars, but generalization to unseen identities is poor, reflected in low CSIM scores.

\paragraph{Effect of {\flextokens}.}
Simply combining monocular and multi-view data (\textit{w/o {\flextokens}}) does not produce complete 3D heads for unseen images. The model mainly learns to identify the dataset rather than resolve viewpoint–expression entanglement. Our full architecture (\textit{Ours\textsuperscript{ref}}) successfully generates complete 3D heads.

\paragraph{Effect of StyleGAN-PixelShuffle upsampler.}
Replacing the StyleGAN-PixelShuffle block with standard PixelShuffle slightly reduces metrics and visual quality, particularly in sensitive facial regions like the eyes and mouth interior.

\paragraph{Effect of Fitting.}
A single forward pass already produces accurate avatars, but fitting further improves identity (CSIM), sharpness (LPIPS), and expression fidelity (AKD). Fitting is fast ($\sim$1 minute) as it optimizes only the avatar code $\mathcal{A}$ while keeping the network frozen.

\begin{figure}
    \centering
    \includegraphics[width=\linewidth]{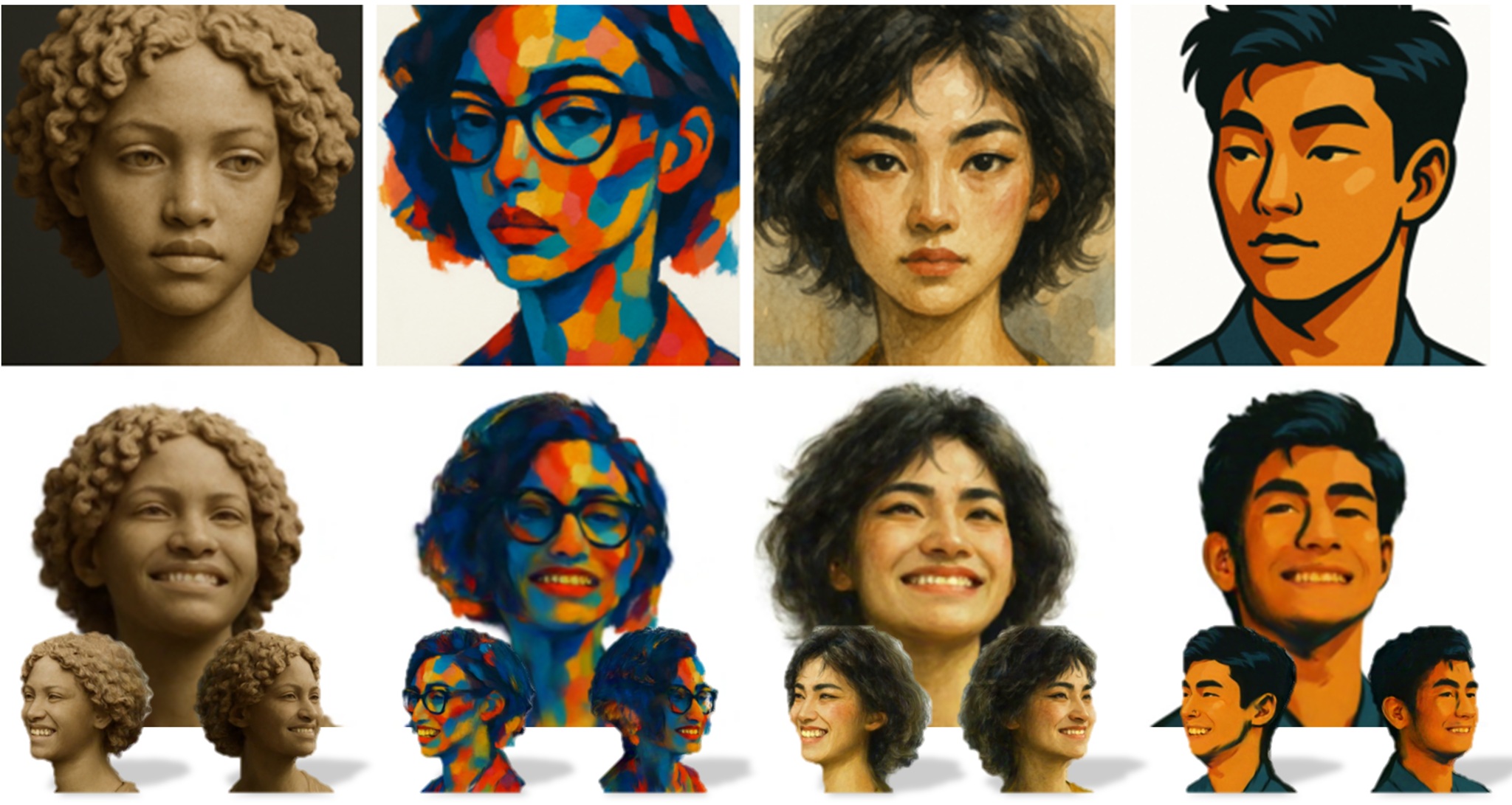}
    \caption{\textbf{In-the-wild results.} We test {\Ours} on highly diverse inputs and perform cross-reenactment.}
    \label{fig:in_the_wild_results}
\end{figure}

\subsection{Limitations.}
While {\Ours} generates high-quality and complete 3D head avatars from a single image, several limitations remain. First, lighting is baked from the input image, preventing explicit control. This can appear unnatural if the avatar is placed in a different virtual environment. Second, although the architecture is 3DMM-free, all experiments use FLAME expression codes, which limits fine details such as the tongue. However, thanks to its model-agnostic design, {\Ours} can be trained with more expressive descriptors, e.g., expression codes from implicit morphable models~\cite{giebenhain2023nphm, potamias2025imhead} or features from generalized expression encoders~\cite{xu2024gphm, tran2024voodooxp, xu2024vasa1}.

\section{Conclusion}

We introduced {\Ours}, a method for generating high-quality, complete 3D head avatars from a single image. Existing methods struggle with view extrapolation. We identify the entanglement between driving signal and target viewpoint in monocular training to be a key issue. To address this, we propose {\flextokens}, which combine the generalization of monocular datasets with the 3D completeness of multi-view supervision. Extensive experiments show that {\Ours} generalizes well and produces realistic avatars. Its smooth latent space enables flexible applications, including few-shot avatar creation from phone scans or monocular videos. 
Our proposed design is quite general and makes little domain-specific assumptions. Extending it to different domains such as human bodies or generalized dynamic novel view synthesis is a promising research direction. Furthermore, we believe our findings on {\flextokens} may benefit other domains where scarce multi-view or 3D data has to be combined with partial supervision from monocular data.

\subsection*{Acknowledgements}
This work was supported by the ERC Consolidator Grant Gen3D (101171131).
We would also like to thank Angela Dai for the video voice-over and Karla Weighart for proof-reading.

{
    \small
    \bibliographystyle{ieeenat_fullname}
    \bibliography{main}
}

\clearpage
\setcounter{page}{1}
\setcounter{table}{5} %
\setcounter{figure}{8} %
\maketitlesupplementary

\appendix

\begin{figure}
    \centering
    \includegraphics[width=\linewidth]{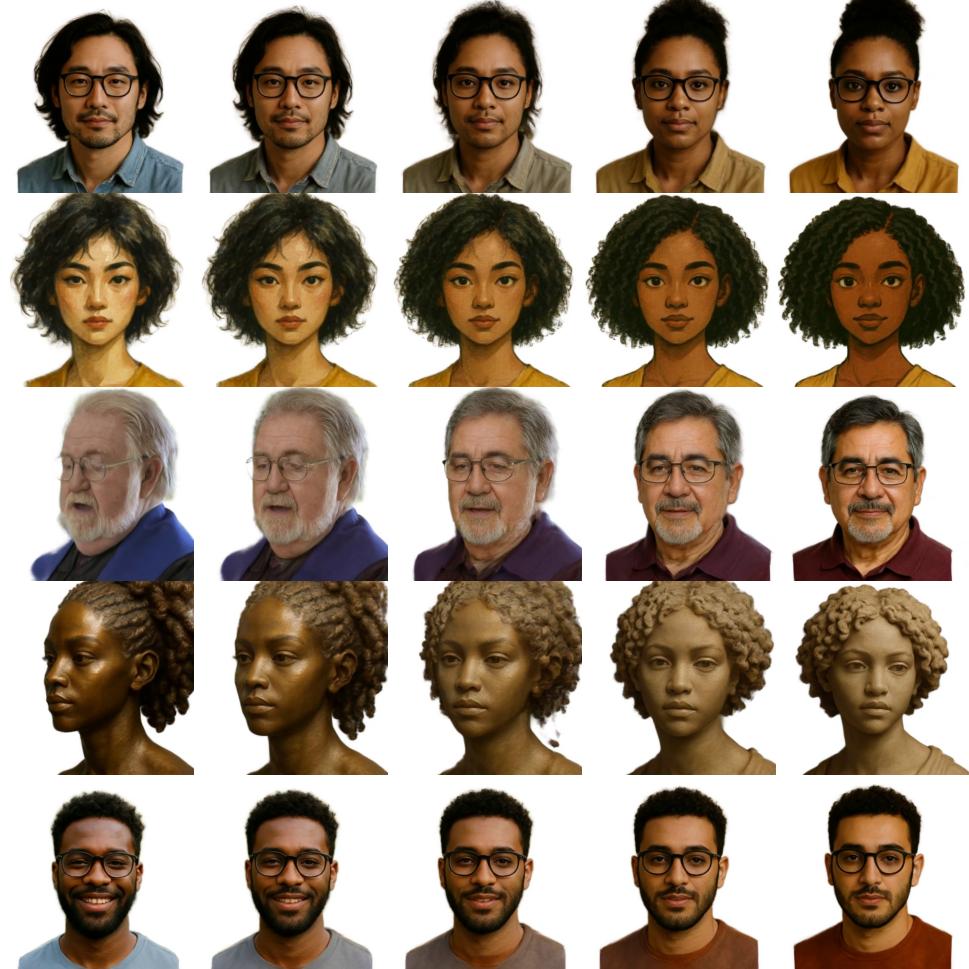}
    \caption{\textbf{Interpolation of 3D Head Avatars.} {\Ours} can produce realistic 3D interpolations between people by interpolating the latent avatar code $\mathcal{A}$, the expression code $z_{exp}$, and the camera $\pi$ of two persons.}
    \label{fig:4_interpolation}
\end{figure}

In this supplementary document, we provide additional comparisons, analysis, and training details. We also highly recommend readers to watch the supplementary video which highlights several aspects of our method, shows plenty of avatars in motion, and features a real-time where a user is walked through the process of creating their own avatar. 

\section{Additional Comparisons}
\subsection{Qualitative Comparison on Portrait Animation}
\cref{fig:x_vfhq_comparison} shows qualitative comparisons on the cross-reenectment setting on the VFHQ test split. We compare with the two most recent baselines GAGAvatar~\cite{chu2024gagavatar} and LAM~\cite{he2025lam}. In both cases, we use the publicly available code to obtain the renderings. Our method produces highly-realistic portrait animations that can capture subtle expressions. Furthermore, our renderings are noticeably sharper than the baselines and contain fewer artifacts, especially under large head rotations of the driver. 

\subsection{Qualitative Comparison on Few-shot Setting}
\begin{figure*}
    \centering
    \includegraphics[width=\linewidth]{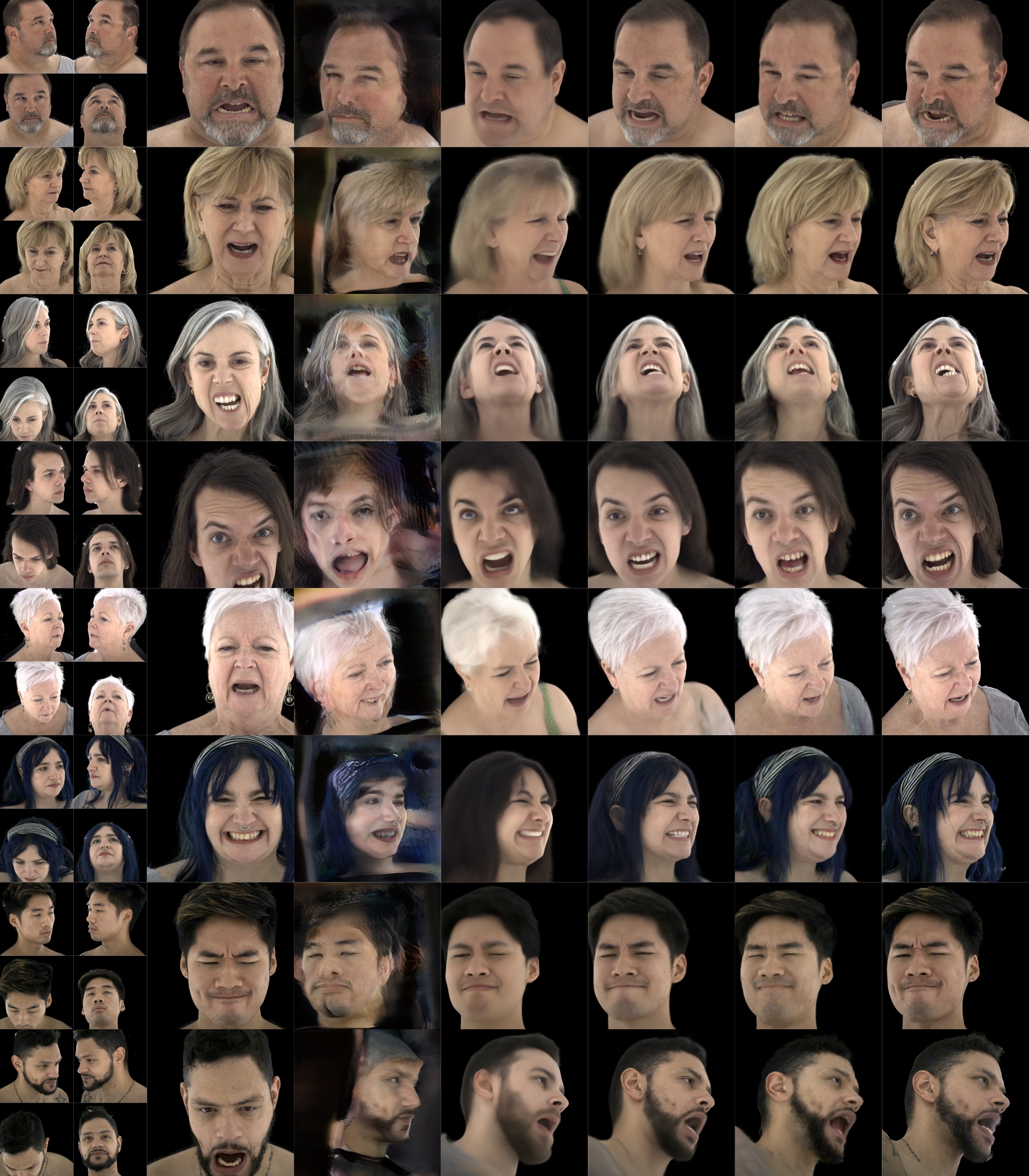}
    \begin{tabularx}{\linewidth}{YYYYYYY}
        Input & Driver & InvertAvatar~\cite{zhao2024invertavatar} & GPAvatar~\cite{chu2024gpavatar} & Avat3r~\cite{kirschstein2025avat3r} & Ours & GT \\
    \end{tabularx}
    \caption{\textbf{Qualitative Few-shot Avatar Creation comparison on the Ava256 dataset.}}
    \label{fig:4_avat3r_comparison}
\end{figure*}

\cref{fig:4_avat3r_comparison} shows qualitative comparisons on the few-shot avatar creation setting following Avat3r~\cite{kirschstein2025avat3r}. Our method creates artifact-free 3D head avatars that closely resemble the input persons and allow expressive animations. 

\begin{figure}[tb]
    \centering
    \includegraphics[width=\linewidth,trim={0 0 0 0},clip]{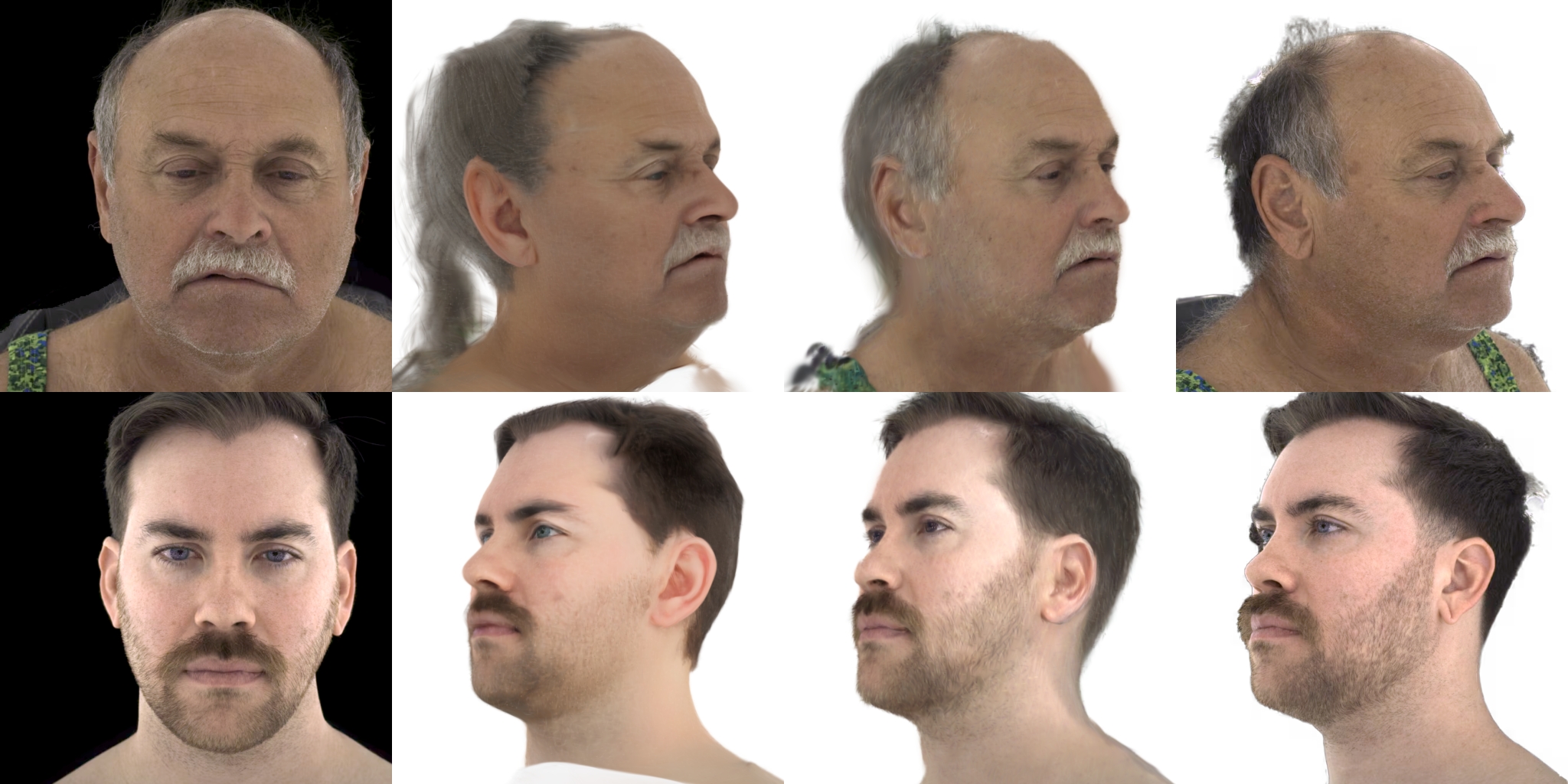}
    \begin{tabularx}{\linewidth}{CCCC}
         Input & FaceLift & Ours & GT
    \end{tabularx}
    \caption{\textbf{Visual comparison with FaceLift on the Ava256 dataset.}}
    \label{fig:facelift_comparison}
\end{figure}

\begin{table}[tb]
    \centering
    \addtolength{\tabcolsep}{-0.4em}
    \resizebox{\linewidth}{!}{
    \begin{tabular}{lrrrrrrrr}
        \toprule
        & \multicolumn{5}{c}{Front} & \multicolumn{3}{c}{Back} \\
        \cmidrule(lr){2-6} \cmidrule(lr){7-9}
        & \small PSNR$\uparrow$ & \small SSIM$\uparrow$ & \small LPIPS$\downarrow$ & \small AKD$\downarrow$ & \small CSIM$\uparrow$ & \small PSNR$\uparrow$ & \small SSIM$\uparrow$ & \small LPIPS$\downarrow$ \\
        \midrule
         FaceLift & 12.8 &  0.715 & 0.357 & 6.32 & 0.658 & 13.2 & 0.687 & 0.411\\
         Ours & \textbf{17.2} & \textbf{0.786} & \textbf{0.265} & \textbf{4.72} & \textbf{0.771} & \textbf{15.2} & \textbf{0.709} & \textbf{0.408} \\
         \bottomrule
    \end{tabular}
    }
    \caption{\textbf{Quantitative comparison with FaceLift on Ava256.}}
    \label{tab:facelift_comparison}
\end{table}

\begin{figure*}[tb]
    \centering
    \setlength{\tabcolsep}{0pt}
    \includegraphics[width=\linewidth,trim={0 0 0 0},clip]{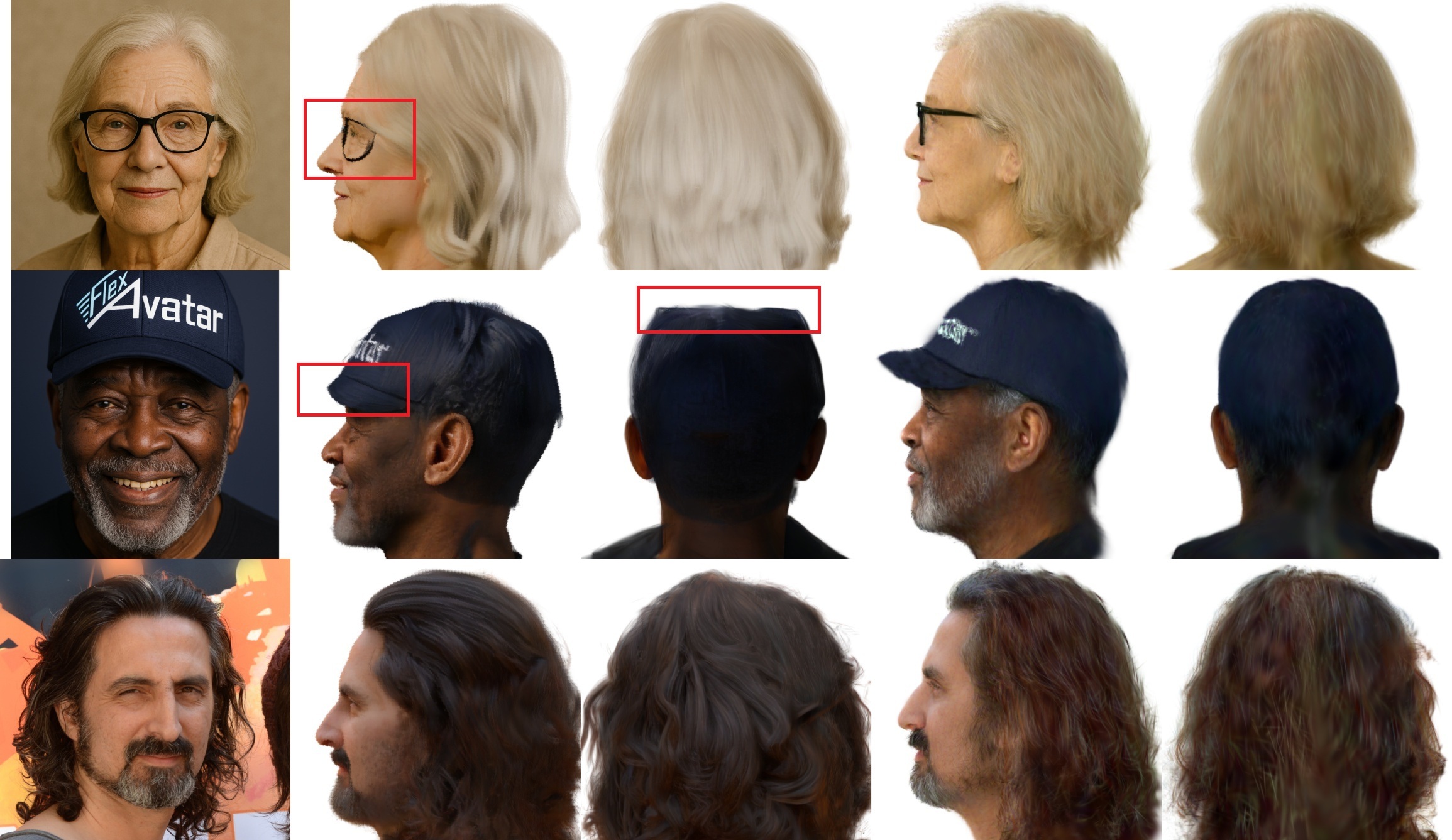}
    \begin{tabularx}{\linewidth}{P{0.2\linewidth}P{0.4\linewidth}P{0.4\linewidth}}
        Input & FaceLift & Ours
    \end{tabularx}
    \caption{\textbf{In-the-wild 360° Comparison with FaceLift.}}
    \label{fig:facelift_itw_comparison}
\end{figure*}

\subsection{Comparison with FaceLift}
We compare our method with FaceLift~\cite{FaceLift} for single-image 3D head reconstruction in two settings: 
\begin{enumerate}
    \item[(i)] On Ava256 (\cref{fig:facelift_comparison} and \cref{tab:facelift_comparison}), we use 4 frontal and 4 back cameras for 5 subjects. Our model slightly outperforms FaceLift quantitatively on back-head renderings and produces noticeably more accurate frontal reconstructions. For fairness, we use a version of our model not trained on Ava256. 
    \item[(ii)] On in-the-wild images (\cref{fig:facelift_itw_comparison}), our method matches FaceLift in completeness while better handling accessories such as caps and glasses. 
\end{enumerate}
In contrast to FaceLift, our method also supports head animation, which is part of our core contribution.

\vspace{0.5cm}
\section{Additional Analyses}
\subsection{Interpolation Between Persons}
Due to the smooth nature of our avatar latent space, we can produce interpolations between persons. This is done by first obtaining the avatar codes from each portrait and then computing a convex combination between them:
\begin{align}
    \mathcal{A}_1 &= E(I_1) \\
    \mathcal{A}_2 &= E(I_2) \\
    \mathcal{A}_{int} &= \alpha \mathcal{A}_1 + (1 - \alpha) \mathcal{A}_2
\end{align}
\cref{fig:4_interpolation} shows example interpolations.

\begin{figure*}[tb]
    \centering
    \includegraphics[width=\linewidth]{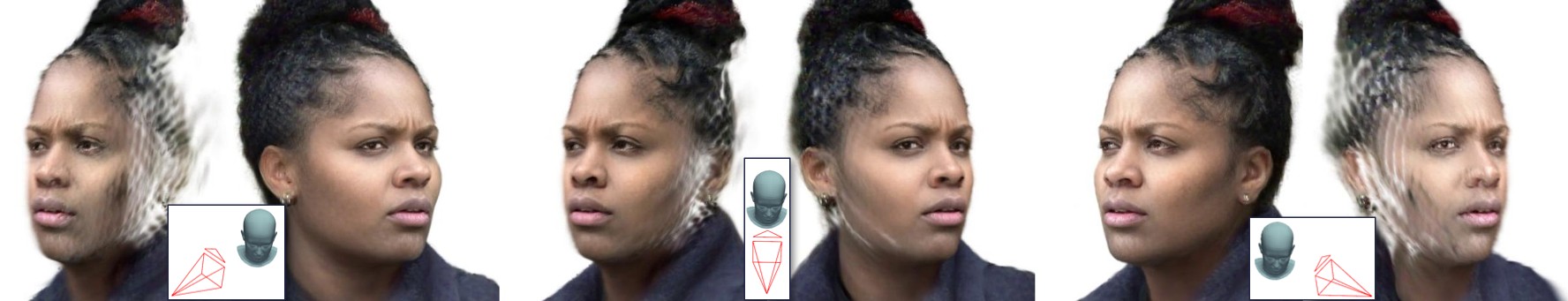}
    \begin{tabularx}{\linewidth}{CCC}
        $z_{bias} = \mathrm{left\,cam}$ & 
        \mbox{$z_{bias} = \mathrm{front\,cam}$}
        & 
        \mbox{$z_{bias} = \mathrm{right\,cam}$}
    \end{tabularx}
    \caption{\textbf{Analysis of Bias Sinks.} In this ablation experiment, there is one bias sink for each of the 16 cameras of the NeRSemble~\cite{kirschstein2023nersemble} dataset. Using the bias sink for a left camera during inference results in a head that is only complete when seen from the left side. Analogously, the bias sink for a right camera leads to the opposite effect. As such, each bias sink effectively captures the viewpoint bias of its respective training data subset. In our full method, we exploit this behavior to obtain a bias sink that produces full 360° heads. }
    \label{fig:r_bias_sinks}
\end{figure*}

\subsection{Analysis of Bias Sinks}
To better understand the effect of bias sinks on the model, we finetune a 2D-only model on the NeRSemble dataset using 1 bias sink per each of the dataset’s 16 cameras. As shown in \cref{fig:r_bias_sinks}, the model learns that the presence of the “left cam” bias sink correlates with a head that is only complete from the left side. This validates, that the bias sinks are an effective way to make the model mirror the behavior of a specific training data subset during inference without loosing generality.

\begin{figure*}[tb]
    \centering
    \includegraphics[width=\linewidth,trim={0 2cm 0 0.5cm},clip]{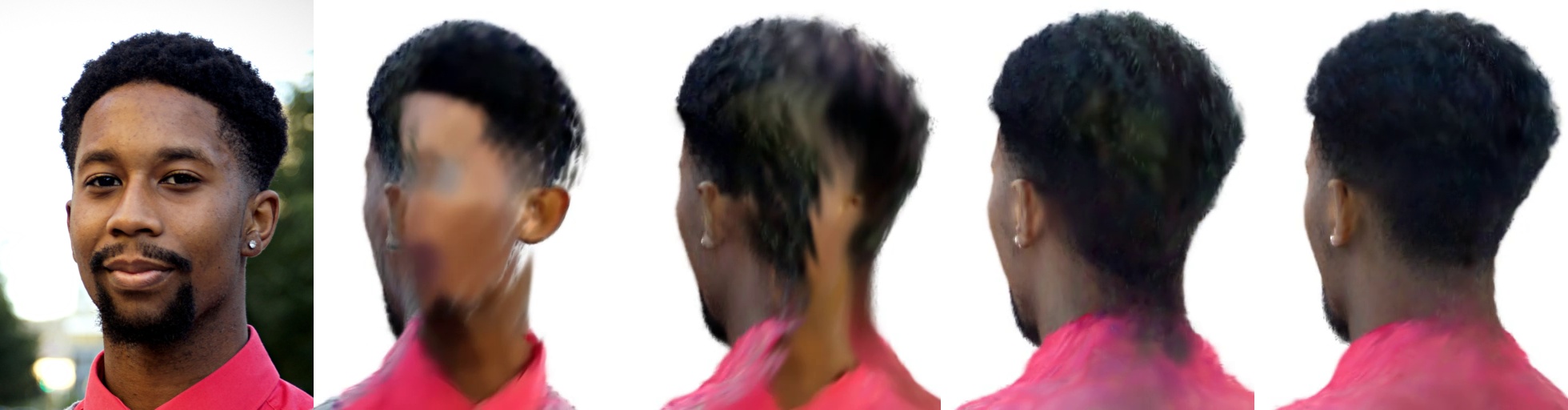}
    \setlength{\tabcolsep}{0pt}
    \begin{tabularx}{\linewidth}{CCCCC}
        Input & Only 2D &  + 1\% 3D &  + 10\% 3D &  + 100\% 3D
    \end{tabularx}
    \caption{\textbf{Amount of 3D data required for bias sinks.} The bias sink mechanism leads to noticeably more complete heads even when only a small proportion of multi-view training data is available.}
    \vspace{-0.3cm}
    \label{fig:r_3d_data_ratio_ablation}
\end{figure*}

\subsection{Analysis of 3D Data Ratio for Bias Sinks}
We analyze how much 3D data is required for the bias sink mechanism to work. To do that, we finetune a 2D-only model with various amounts of multi-view data. \cref{fig:r_3d_data_ratio_ablation} shows that the bias sinks already lead to noticeably more complete heads with only 1\% of the 3D training data (=17 different people). Gradually increasing the amount of 3D training data makes the bias sinks more effective with 10\% (=186 people) already producing a complete 3D head.

\begin{figure*}[tb]
    \centering
    \includegraphics[width=\linewidth]{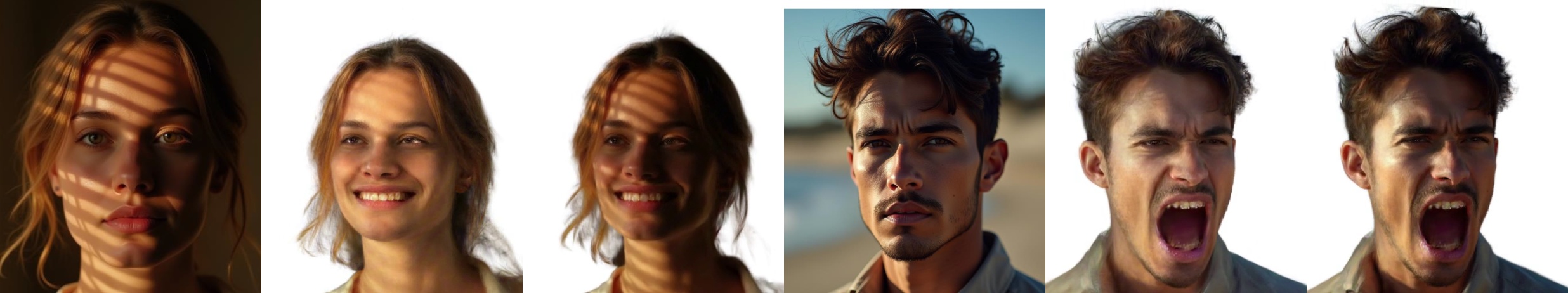}
    \setlength{\tabcolsep}{0pt}
    \begin{tabularx}{\linewidth}{CCCCCC}
        Input & \mbox{w/o fitting} & Ours & Input & \mbox{w/o fitting} & Ours
    \end{tabularx}
    \caption{\textbf{Performance under challenging lighting.} Example inputs taken from IC-Light~\cite{zhang2025iclight}.}
    \label{fig:r_ic_light}
\end{figure*}

\subsection{Analysis of Robustness}
In~\cref{fig:r_ic_light}, we show our method on 2 challenging lighting situations. Sole inference with $z_{3D}$ (w/o fitting) attenuates shadows due to the even lighting bias of multi-view data. This is resolved in our full pipeline with fitting. We also refer to our supplementary video that contains 57 avatars from in-the-wild images. 

\subsection{FPS and VRAM usage}
During inference, our model needs 1.7GB of VRAM. Animation and rendering run at 20 fps on an RTX 3090 GPU. Avatar creation, including all processing, takes 2 minutes. For a demonstration, see the live demo in the supplemental video. 

\begin{figure}[tb]
    \centering
    \includegraphics[width=\linewidth,trim={0 0 0 1cm},clip]{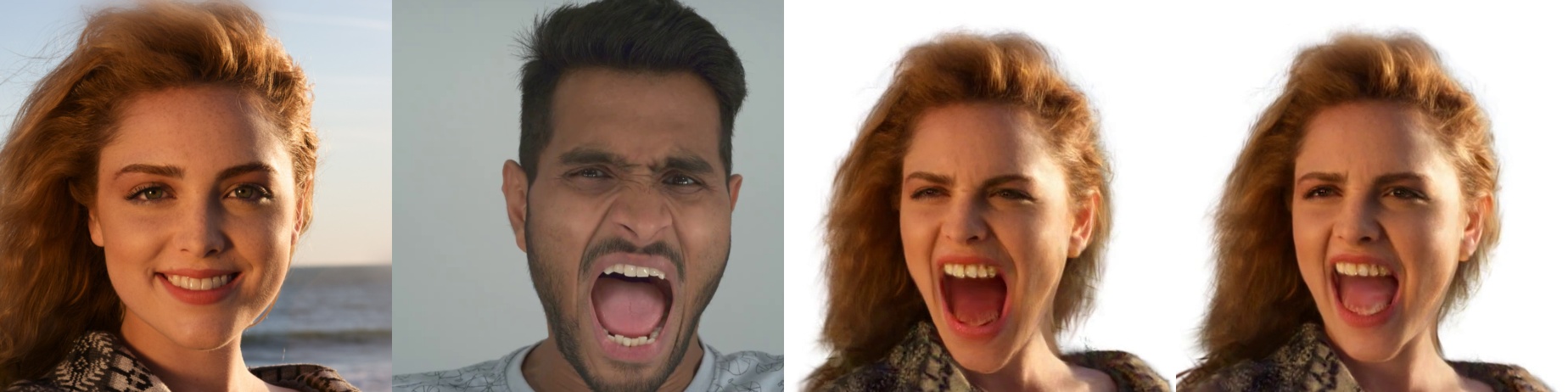}
    \begin{tabularx}{\linewidth}{CCCC}
        \small{Input} & \small{Driver} & \small{FLAME} & \small{VOODOOXP}
    \end{tabularx}
    \caption{\textbf{Analysis of FLAME dependence.} FlexAvatar can be trained with different expression control signals such as expression codes from VOODOOXP~\cite{tran2024voodooxp}.}
    \label{fig:r_voodooxp_ablation}
\end{figure}

\subsection{Analysis of FLAME dependence}
We finetune our model using codes from VOODOOXP's expression encoder~\cite{tran2024voodooxp} instead of FLAME expression codes. \cref{fig:r_voodooxp_ablation} shows that our method is not dependent on FLAME's expression space.

\begin{figure}[t]
    \centering
    \includegraphics[width=\linewidth]{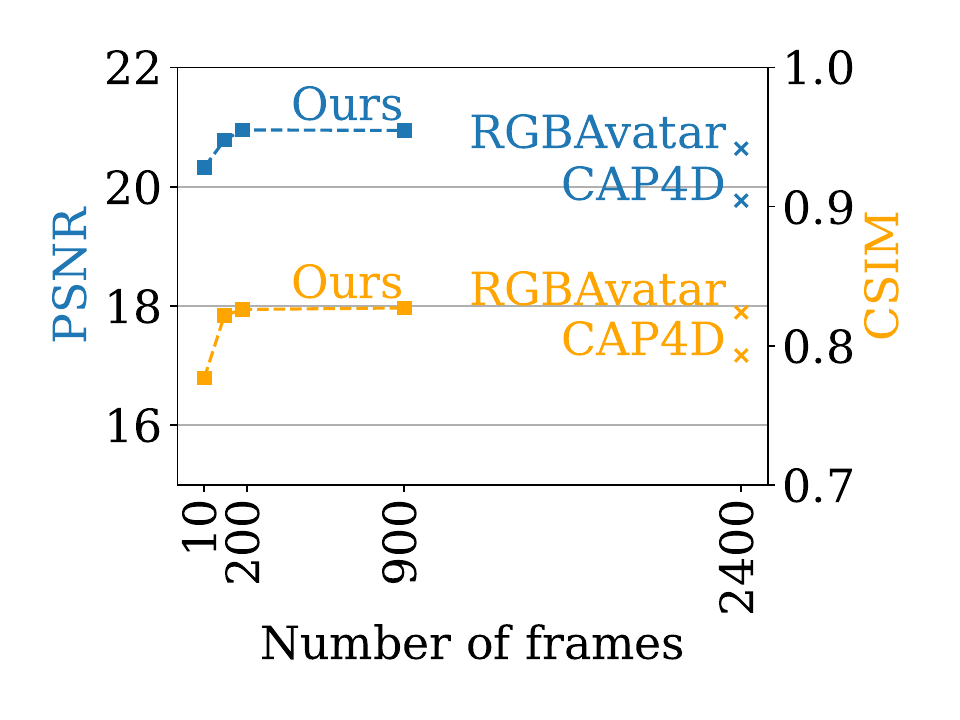}
    \caption{\textbf{Analysis of Data Efficiency during fitting.} We plot the performance of our method on the NeRSemble Benchmark~\cite{kirschstein2023nersemble} in relation to how many frames of a person were used during fitting to create the avatar. Note that the two most competitive baselines on the benchmark, RGBAvatar~\cite{li2025rgbavatar} and CAP4D~\cite{taubner2025cap4d} use all available frames while our method requires only $\sim\frac{1}{10}$ of the frames for a competitive performance. By using $\sim\frac{2}{5}$ of the frames, {\Ours} outperforms the baselines.}
    \label{fig:x_nersemble_data_efficiency}
\end{figure}

\subsection{Analysis of Data Efficiency during Fitting}
In~\cref{fig:x_nersemble_data_efficiency}, we analyze how the quality of an avatar increases with the number of available input images. To do so, we use the monocular videos from the 5 NeRSemble benchmark~\cite{kirschstein2023nersemble} persons and apply the fitting procedure as described in the main paper with 2000 optimization steps. It can be seen that both image quality (PSNR) as well as identity preservation (CSIM) greatly increase with the first $\sim$100 frames and level off after that. We achieve competitive performance on the benchmark with an order of magnitude less input frames required.

\section{Training Details}
\subsection{Data Preparation}
To remove the background in the training videos, we use MatAnyone~\cite{yang2025matanyone}. For single input images during inference, we use MODNet~\cite{ke2022modnet}. We also use MODNet to segment out the background in the generations of GAGAvatar~\cite{chu2024gagavatar} in the supplemental video and in~\cref{fig:x_vfhq_comparison}. This is because GAGAvatar can only render images with black background due to its use of a screen-space renderer.

\paragraph{Head-centric coordinates.}
We simplify the models task by always predicting the avatar in FLAME's canonical space, i.e., factoring out the effect of rigid head movement. To do this, the rigid head transformation matrix is instead applied to the cameras. During inference, head movement is then also modelled by factoring the head motion into the rendering viewpoint. As a side effect, it becomes harder for the model to predict the correct torso pose which has to move relative to the canonical head pose. 

\paragraph{Expression codes.}
As it can be seen in~\cref{fig:r_voodooxp_ablation}, our architecture is agnostic to the specific choice of animation signal. In our experiments, we use FLAME expression codes obtained from Pixel3DMM~\cite{giebenhain2025pixel3dmm}. However, note that our design allows to train on different animation signals without any change to the architecture itself. Possible animation controls may be expression codes from implicit morphable head models~\cite{giebenhain2023nphm} or codes derived from speech.

\begin{table}[tb]
    \centering
    \resizebox{\linewidth}{!}{
    \begin{tabular}{lll}
        \toprule
        & Hyperparameter & Value \\
        \midrule
            \multirow{6}{*}{\rotatebox[origin=c]{90}{\parbox[c]{2cm}{\centering \small Architecture}}} 
                & ViT patch size & $16 \times 16$ \\
                & hidden dimension $D$ & 768 \\
                & \#cross-attention layers in encoder & 8 \\
                & \#cross-attention layers in decoder & 8 \\
                & \#StyleGAN-PixelShuffle layers & 2 \\
                & Size of avatar code $\mathcal{A}$ & $32 \times 32 \times 768$ \\
         \midrule
            \multirow{4}{*}{\rotatebox[origin=c]{90}{\parbox[c]{1.2cm}{\centering \small In \& Out}}} 
                & Input image resolution & $512 \times 512$ \\
                & Train render resolution & $512 \times 512$ \\
                & Gaussian attribute map resolution & $256 \times 256$ \\
                & \#3D Gaussians & $\sim$ 58k \\

         \midrule
            \multirow{3}{*}{\rotatebox[origin=c]{90}{\parbox[c]{1.5cm}{\centering \small Expression MLP}}} 
                & Dimension of expression code & 135 \\
                & \#expression sequence MLP layers & 2 \\
                & Dimension of expression sequence MLP & 256 \\
                & Expression sequence MLP activation & ReLU \\
         \bottomrule
         
    \end{tabular}
    }
    \vspace{-0.3cm}
    \caption{\textbf{Hyperparameters.}}
    \vspace{-0.5cm}
    \label{tab:x_hyperparameters}
\end{table}

\begin{figure*}
    \centering
    \includegraphics[width=\linewidth]{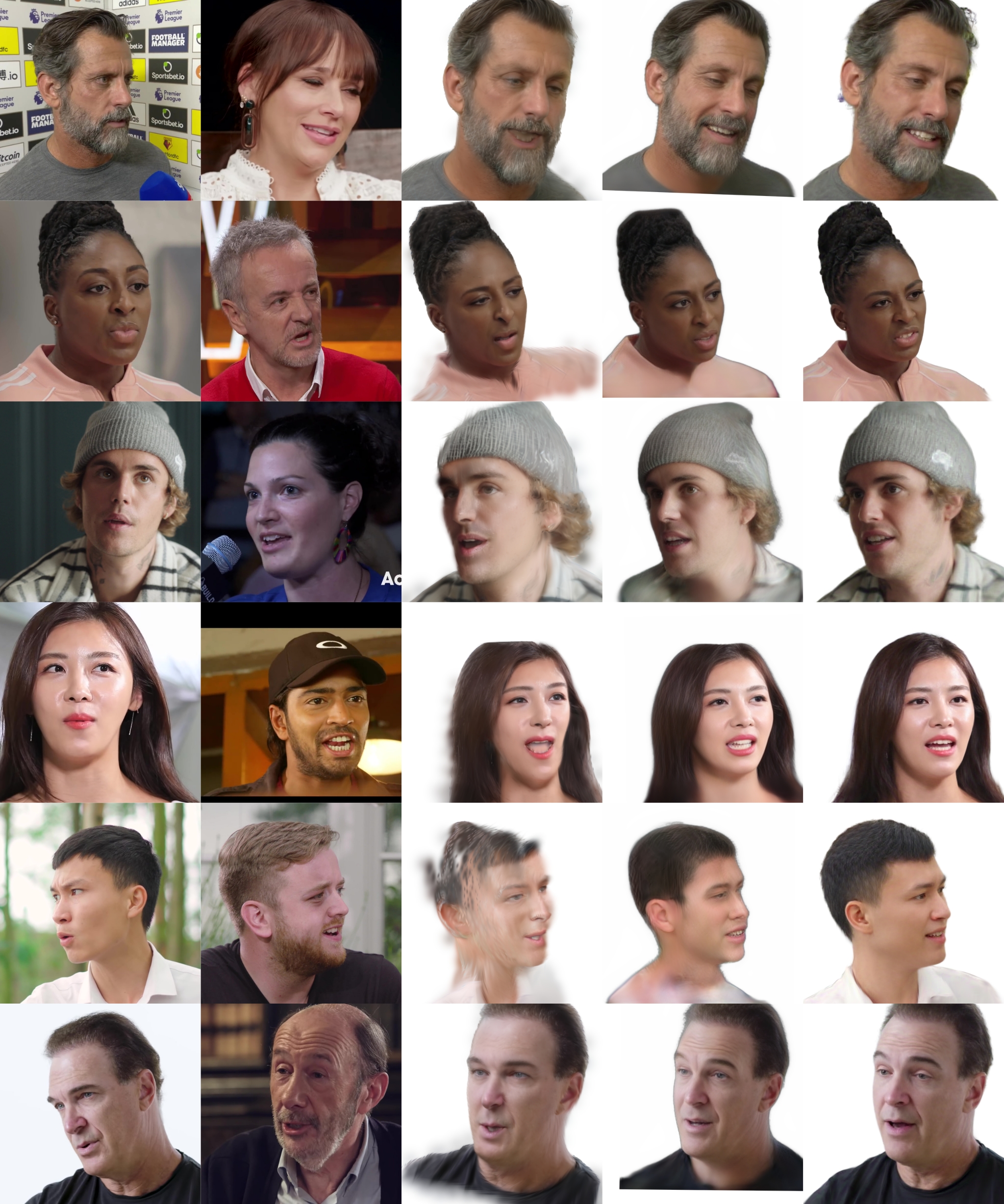}
    \begin{tabularx}{\linewidth}{YYYYY}
        Input & Driver & LAM~\cite{he2025lam} & GAGAvatar~\cite{chu2024gagavatar} & Ours \\
    \end{tabularx}
    \caption{\textbf{Qualitative Portrait Animation with cross-reenactment on the VFHQ test split.}}
    \label{fig:x_vfhq_comparison}
\end{figure*}

\end{document}